\documentclass{article}

\usepackage{microtype}
\usepackage{graphicx}
\usepackage{booktabs}

\usepackage{amsmath}
\usepackage{amsfonts}
\usepackage{subfig}

\usepackage[accepted]{icml2021_oppo}

\usepackage{graphicx}
\newcommand{\subf}[2]{%
  {\small\begin{tabular}[t]{@{}c@{}}
  #1\\#2
  \end{tabular}}%
}

\usepackage{hyperref}
\icmltitlerunning{Implicit Greedy Rank Learning in Autoencoders via Overparameterized Linear Networks}

\begin{document}

\twocolumn[
\icmltitle{Implicit Greedy Rank Learning in Autoencoders \\
via Overparameterized Linear Networks}

\icmlsetsymbol{equal}{*}

\begin{icmlauthorlist}
\icmlauthor{Shih-Yu Sun}{apple}
\icmlauthor{Vimal Thilak}{apple}
\icmlauthor{Etai Littwin}{apple}
\icmlauthor{Omid Saremi}{apple}
\icmlauthor{Joshua M. Susskind}{apple}
\end{icmlauthorlist}

\icmlaffiliation{apple}{Apple}
\icmlcorrespondingauthor{Joshua M. Susskind}{jsusskind@apple.com}
\icmlkeywords{overparameterization, greedy, low-rank, autoencoder}

\vskip 0.3in
]

\printAffiliationsAndNotice{}

\begin{abstract}
Deep linear networks trained with gradient descent yield low rank solutions, as is typically studied in matrix factorization. In this paper, we take a step further and analyze implicit rank regularization in autoencoders. We show greedy learning of low-rank latent codes induced by a linear sub-network at the autoencoder bottleneck. We further propose orthogonal initialization and principled learning rate adjustment to mitigate sensitivity of training dynamics to spectral prior and linear depth. With linear autoencoders on synthetic data, our method converges stably to ground-truth latent code rank. With nonlinear autoencoders, our method converges to latent ranks optimal for downstream classification and image sampling.
\end{abstract}

\section{Introduction}
Gradient-based optimization is biased towards low-rank solutions in deep matrix factorization, which is enhanced by increased depth \citep{deep_matrix_fac, bias_rank, greedy}. Based on insights from matrix factorization, recent works empirically studied low-rank bias in nonlinear deep networks \citep{irmae,low-rank-simplicity-bias,expandnets}. In particular, \citep{irmae} empirically show that latent code rank can be implicitly regularized by using an overparameterized linear network at the bottleneck of an autoencoder (AE). The learned low-rank codes have competitive quality to those from more sophisticated encoders like VAE \citep{VAE}. However, linear sub-network depth needs to be tuned by additional downstream tasks to avoid rank collapse and catastrophic performance due to too large depth, as shown in Figure~\ref{fig:std_init_nonlinear}.

Motivated by \citep{irmae}, in this paper we study gradient descent dynamics of an overparameterized \footnote{Overparameterization could refer to excessive parameters against data samples or function space. Our usage here is more aligned with the latter as in \citep{deep_linear_acceleration}.} linear sub-network at the AE bottleneck, building on theoretical results on the acceleration effect from \citep{deep_linear_acceleration}, with the aim of estimating latent code ranks that reflect intrinsic data dimensions in an unsupervised setting. Our contributions are summarized as follows:
\vspace{-7pt}
\begin{itemize}
\setlength\itemsep{0em}
\item We analyze training dynamics of a linear sub-network in an AE and show greedy learning of latent rank, and reveal spectral bias induced by standard initialization, which we address by scaled orthogonal initialization. 
\item We further propose principled learning rate adjustment to stabilize the greedy-learning dynamics with respect to linear sub-network depth.
\item We validate our method for linear and nonlinear AE, where it converges to ground-truth synthetic data rank and a range of ranks optimal for downstream tasks, without task hyperparameter tuning.
\end{itemize}

\section{Implicit Greedy Rank Learning}
We study training dynamics of a linear sub-network inserted at the bottleneck of an AE, which reveals the behavior of greedy rank learning. Given an encoder $enc(\cdot)$ that transforms input $\bold{x} \in \mathbb{R}^c$ into a latent code $\bold{z} \in \mathbb{R}^d$, a sub-network of $N$ linear layers with weights $W_i \in \mathbb{R}^{d \times d}$ further transforms the latent code, which is then fed to a decoder $dec(\cdot)$ to generate reconstruction $\bold{x'} \in \mathbb{R}^c$:
\begin{align*}
\bold{z} &= enc(\bold{x}) \\
\bold{z}_i &= 
\begin{cases}
\bold{z} & \text{for } i=0\\
W_i \bold{z}_{i-1} & \text{for } i = 1,\cdots,N\\
\end{cases} \\
\bold{x'} &= dec(\bold{z}_N)
\end{align*}
Hence, $\bold{z}_N = W_\text{e} \bold{z}$, where $W_\text{e}$ denotes the effective matrix: $W_\text{e} = W_N W_{N-1} \dots W_2 W_1$. A loss function $L(\bold{x}, \bold{x'})$ is defined to measure dissimilarity between $\bold{x}$ and $\bold{x'}$ within a batch of samples (e.g. MSE loss). The set of weights $\{W_i\}$ is jointly optimized with $enc(\cdot)$ and $dec(\cdot)$ by minimizing the loss with gradient descent, and serves as a rank regularizer of $\bold{z}_N$. For inference, $\{W_i\}$ collapses to a single matrix or is absorbed into adjacent linear layers, so little to no overhead is introduced.

\subsection{Updates of Effective Matrix $W_\text{e}$}
We examine gradient descent of $W_\text{e}(t)$ by applying main results from \citep{deep_linear_acceleration}, which state the following. Assume balanced initialization, defined as:
\begin{align}
\label{eq:init_assumption}
    W_{i+1}^\top(0) W_{i+1}(0) = W_i(0) W_i^\top(0),
\end{align} which can be satisfied with near-zero weights.
Given learning rate $\eta$, updates of $W_\text{e}(t)$ (denoted by $\Delta W_\text{e}$) are described by: (dropping $t$ to avoid symbol clutter)
\begin{align} 
\label{eq:dynamic2}
    vec(\Delta W_\text{e}) = -\eta P_{W_\text{e}} vec(\frac{dL}{dW}(W_\text{e}))
\end{align}
$vec(A)$ denotes the vector form of a matrix $A$ in column-first order. $P_{W_\text{e}} \in \mathbb{R}^{d^2 \times d^2}$ is a positive semi-definite matrix that depends on $W_\text{e}$. Denoting SVD of $W_\text{e}$ as $W_\text{e}=U \Sigma V^\top$, eigenvalue decomposition of $P_{W_\text{e}}$ could be written as:
\begin{align}
    P_{W_\text{e}} &= \sum_r \sum_{r'} \mu_{r,r'} \bold{e}_{r,r'} \bold{e}_{r,r'}^\top \label{eq:eigen_decomp} \\
    \bold{e}_{r,r'} &= vec(\bold{u}_r \bold{v}_{r'}^\top) \\
    \mu_{r,r'} &= \sum_{j=1}^N \sigma_{r}^{2\frac{N-j}{N}} \sigma_{r'}^{2\frac{j-1}{N}} \label{eq:mu}
\end{align}
where $\mathbf{u}_r$ and $\mathbf{v}_{r'}$ are columns of $U$ and $V$, respectively, and $\sigma_r$ are singular values of $W_\text{e}$.

In the AE setup, we have:
\begin{align}
\label{eq:loss_gradient}
    \frac{dL}{dW}(W_\text{e}) = \frac{dL}{d\bold{z}_N}^\top \bold{z}^\top
\end{align}
Plugging Equations \ref{eq:eigen_decomp} and \ref{eq:loss_gradient} into Equation \ref{eq:dynamic2}, we have:
\begin{align}
    vec(\Delta W_\text{e}) &= -\eta \sum_r \sum_{r'} \mu_{r,r'} \bold{e}_{r,r'} (\bold{e}_{r,r'}^\top vec(\frac{dL}{d\bold{z}_N}^\top \bold{z}^\top)) \nonumber \\
    &= -\eta \sum_r \sum_{r'} \mu_{r,r'} \langle \bold{u}_r, \frac{dL}{d\bold{z}_N}^\top \rangle \langle \bold{v}_{r'}, \bold{z} \rangle \bold{e}_{r,r'}
    \label{eq:update} 
\end{align}

In the following sections, we discuss implications of Equation~\ref{eq:update} on learning dynamics of $W_\text{e}$.

\subsection{Greedy Rank Learning of $W_\text{e}$}
Equation~\ref{eq:update} reveals $\Delta W_\text{e}$ is dominated by singular vectors of $W_\text{e}$ with stronger presence (i.e. larger $\mu_{r,r'}$) and better alignment with training signals, $\frac{dL}{d\bold{z}_N}$ and $\bold{z}$. This acceleration effect makes the dominant singular vector even more pronounced, forming a self-reinforcing loop. The dominant singular value increases until reconstruction is close to the input along that latent direction, resulting in diminished presence of this direction in $\frac{dL}{d\bold{z}_N}$. Weight updates then turn to focus on the next dominant direction present in training signals. In Appendix~\ref{appendix:greedy}, evolution of top singular values shows this greedy learning behavior.
We hypothesize that new components stop emerging when update directions become less aligned across training steps. Hence, training converges to a rank that captures consistently occurring input characteristics across iterations, and neglects rarely occurring ones. This insensitivity to transient signals mitigates overfitting to noise. Our hypothesis is supported by experimental results as discussed in Section~\ref{sec:experiments}.

\subsection{Orthogonal Initialization of $W_i$}
\label{sec:orthogonal_init}
From Equation~\ref{eq:update}, $\Delta W_\text{e}(t)$ is modulated by spectrum of $P_{W_\text{e}}$ as a function of $W_\text{e}(t)$. This observation reveals downsides of standard initialization for $W_i$ \citep{He_init, Xavier_init}: it leads to a non-uniform random spectrum of $P_{W_\text{e}}$, which could sub-optimally modulate training signals. Further, this spectral prior depends on the linear sub-network depth $N$, whose influence on training dynamics could then be hard to predict and hence needs to be tuned. To decouple spectral structures induced by initialization from those prompted by training signals, we propose use of orthogonal initialization \citep{orthogonal_init} on $W_i$ to initialize spectrum of $W_\text{e}$ uniformly and independently of $N$. Note that in \citep{deep_linear_acceleration}, the initialization condition (Equation \ref{eq:init_assumption}) is satisfied by near-zero weights. It is also satisfied by orthogonal matrices of an arbitrary constant scale, and hence their results are applicable in this case.

\begin{figure*}[t!]
\centering
\begin{tabular}{cccc}
\subf{\includegraphics[width=38mm]{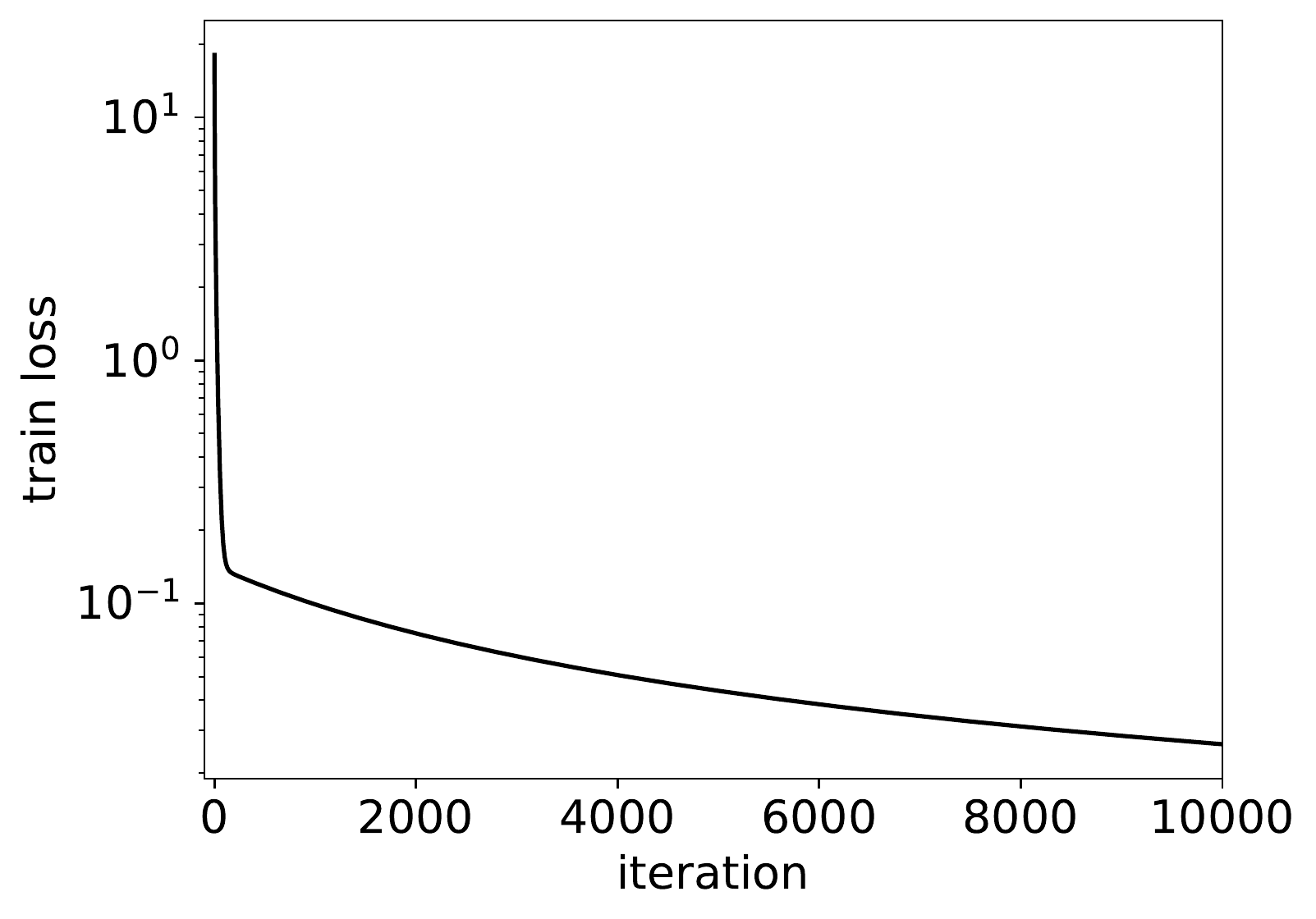}}{} &
\subf{\includegraphics[width=40mm]{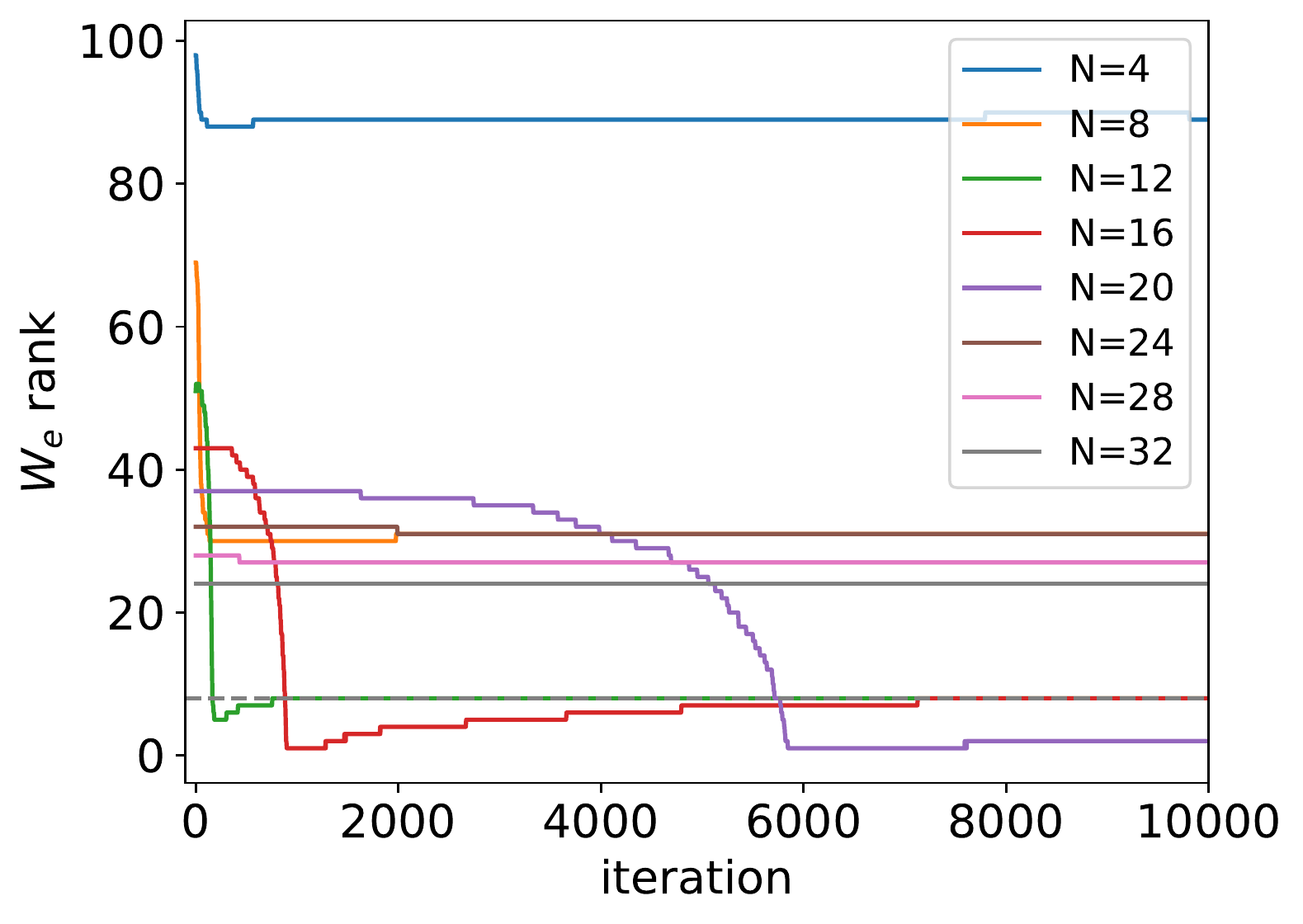}}{} &
\subf{\includegraphics[width=40mm]{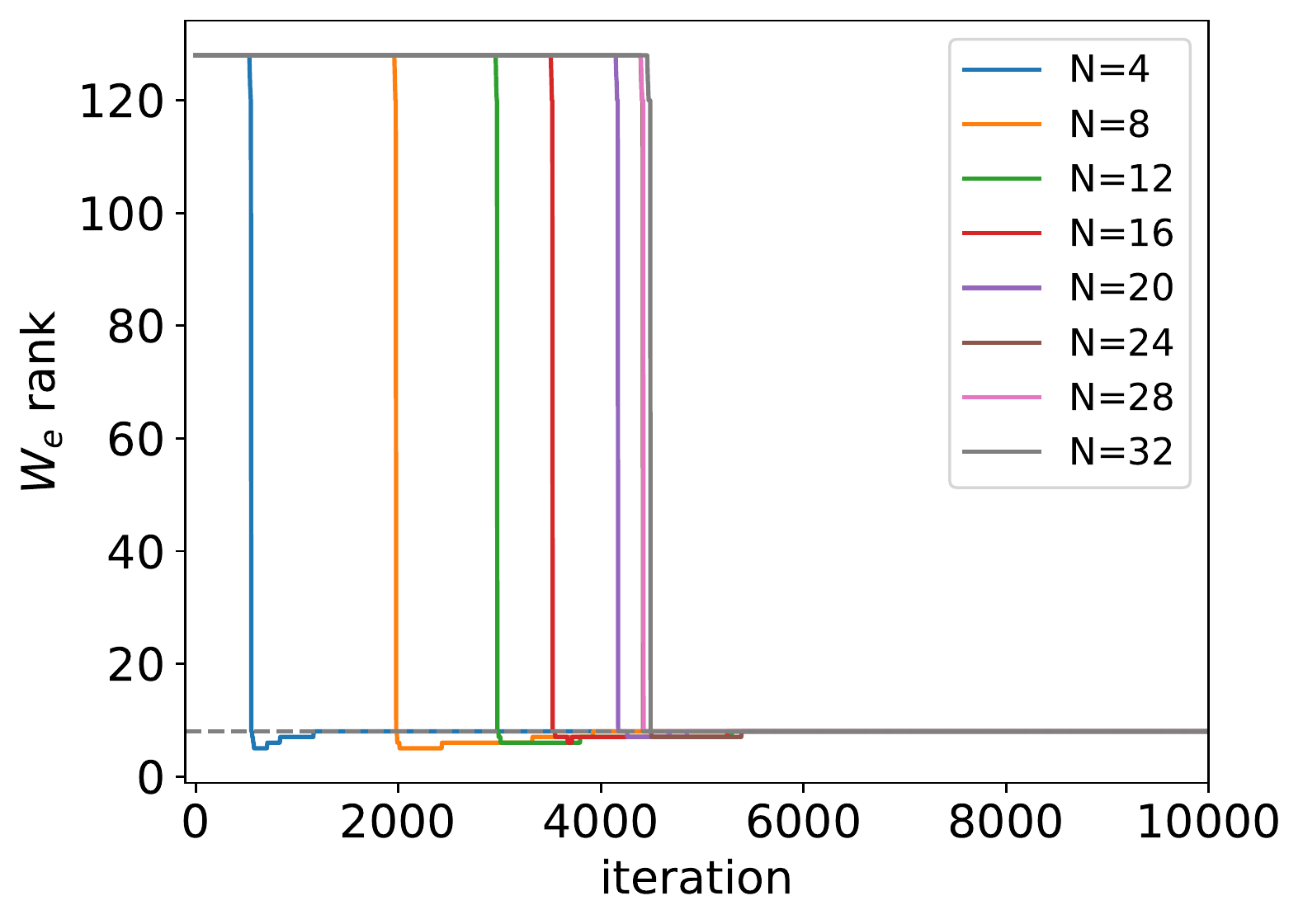}}{} &
\subf{\includegraphics[width=40mm]{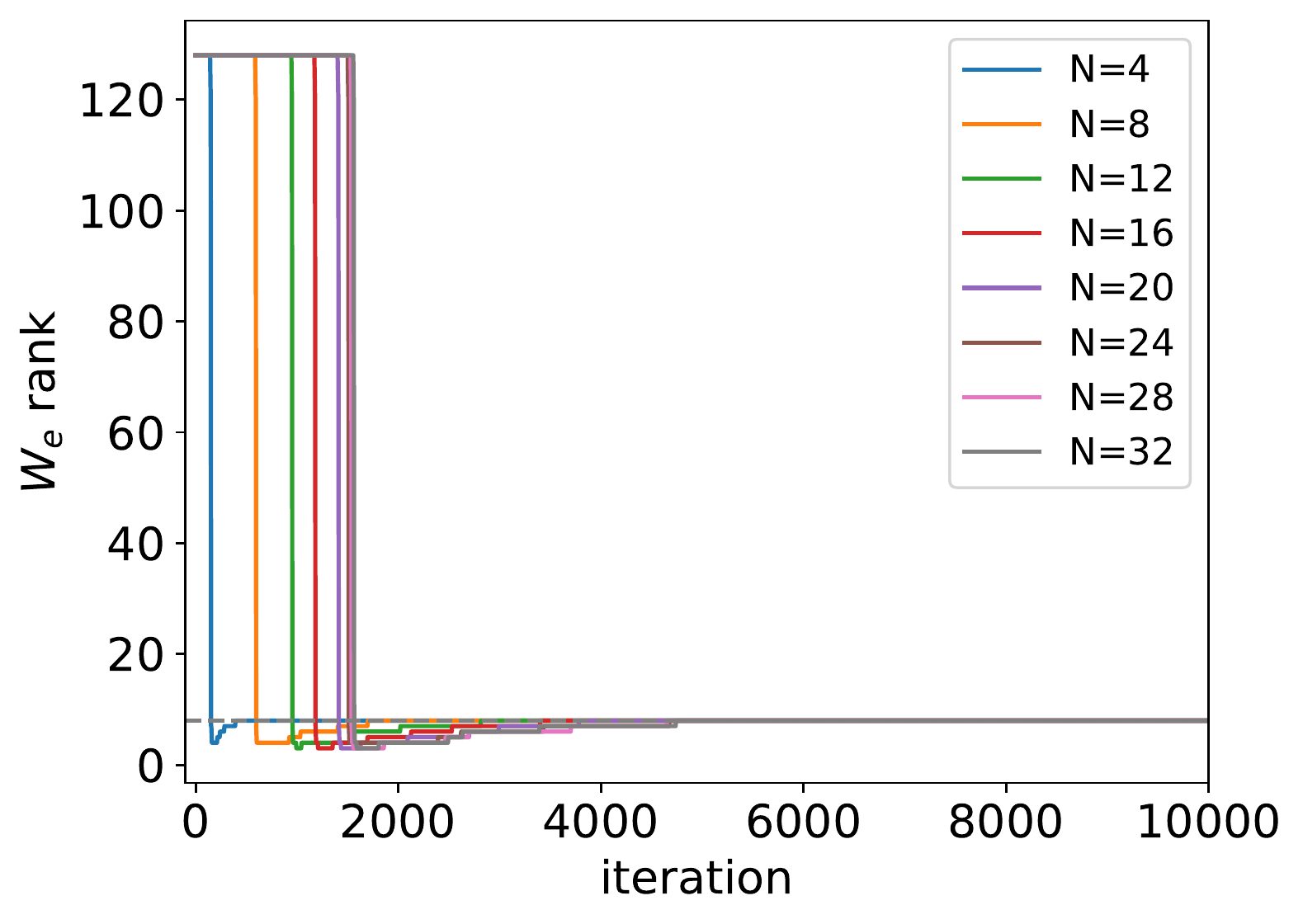}}{}
\\
\subf{\includegraphics[width=38mm]{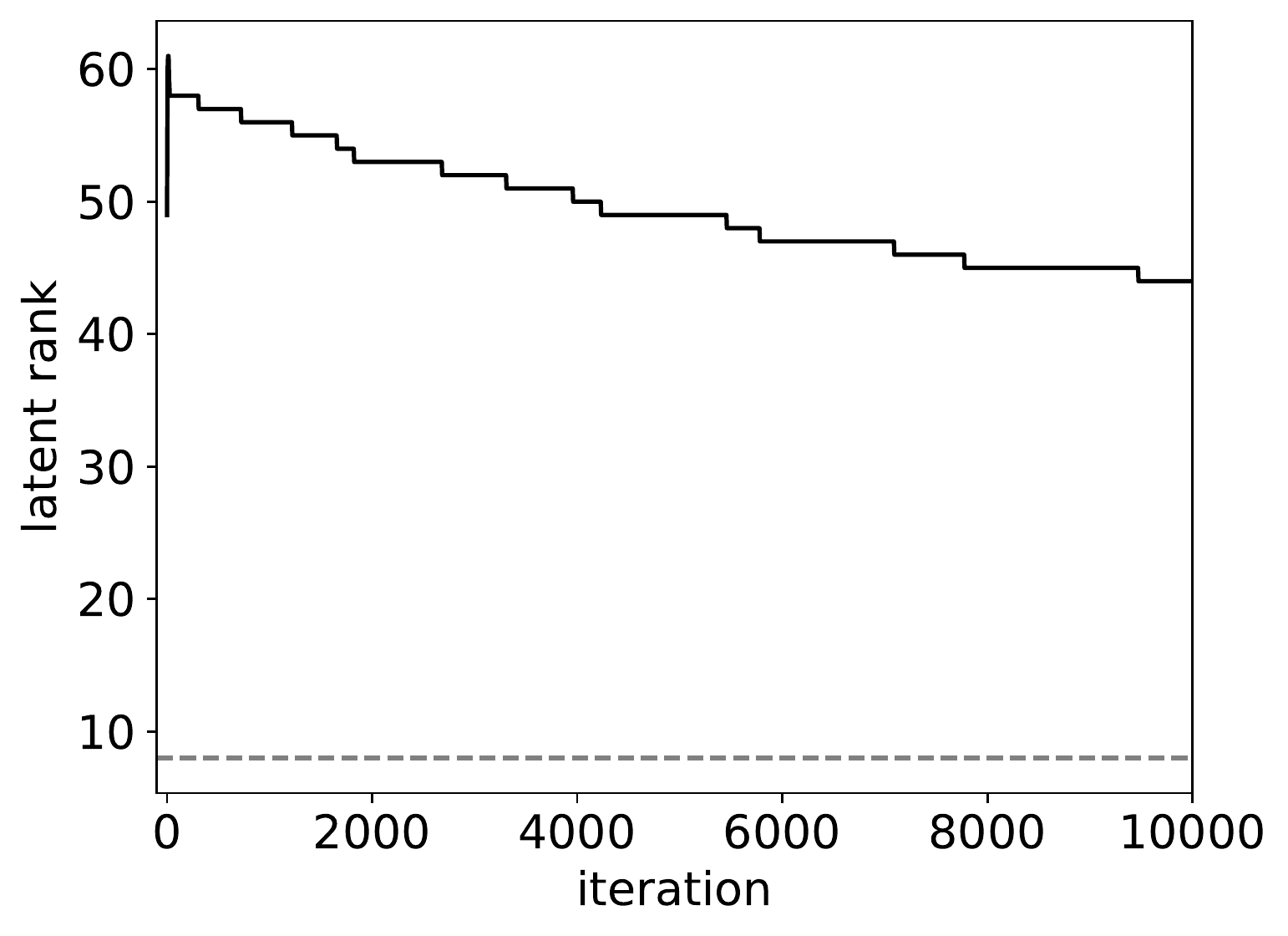}} {(a) vanilla AE} &
\subf{\includegraphics[width=40mm]{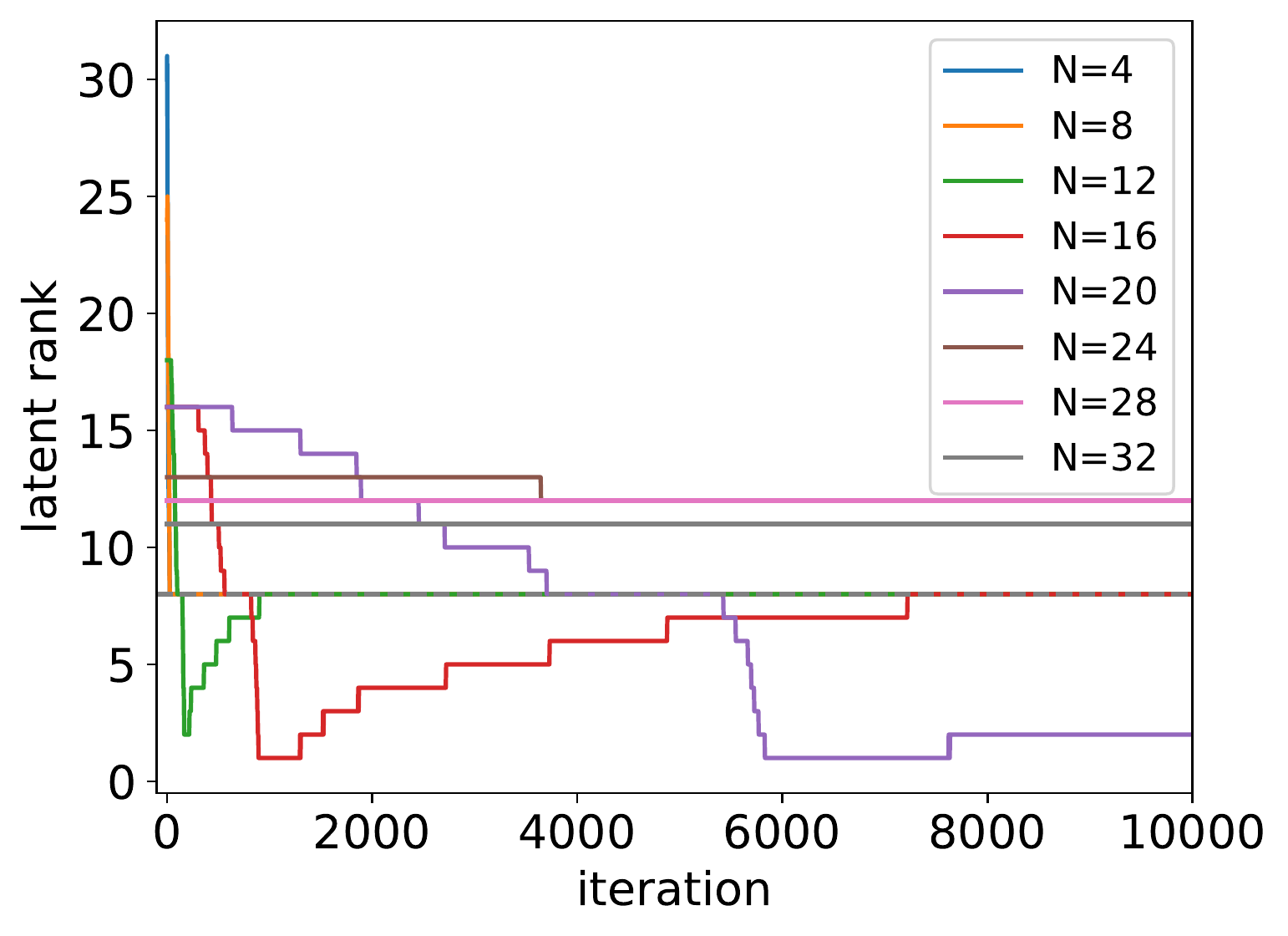}} {(b) standard init} &
\subf{\includegraphics[width=40mm]{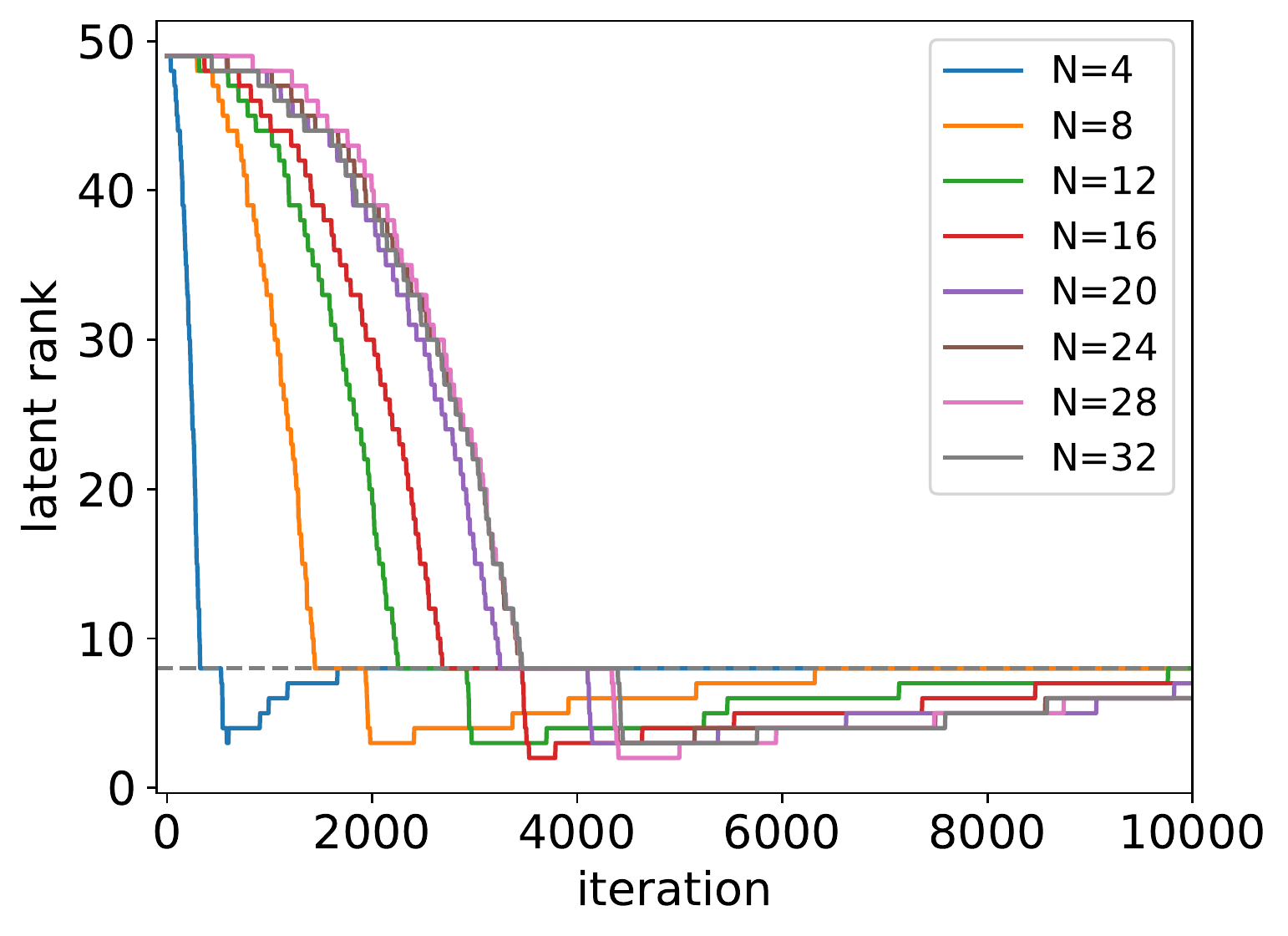}} {(c) orthogonal init ($\alpha=1.0$)} &
\subf{\includegraphics[width=40mm]{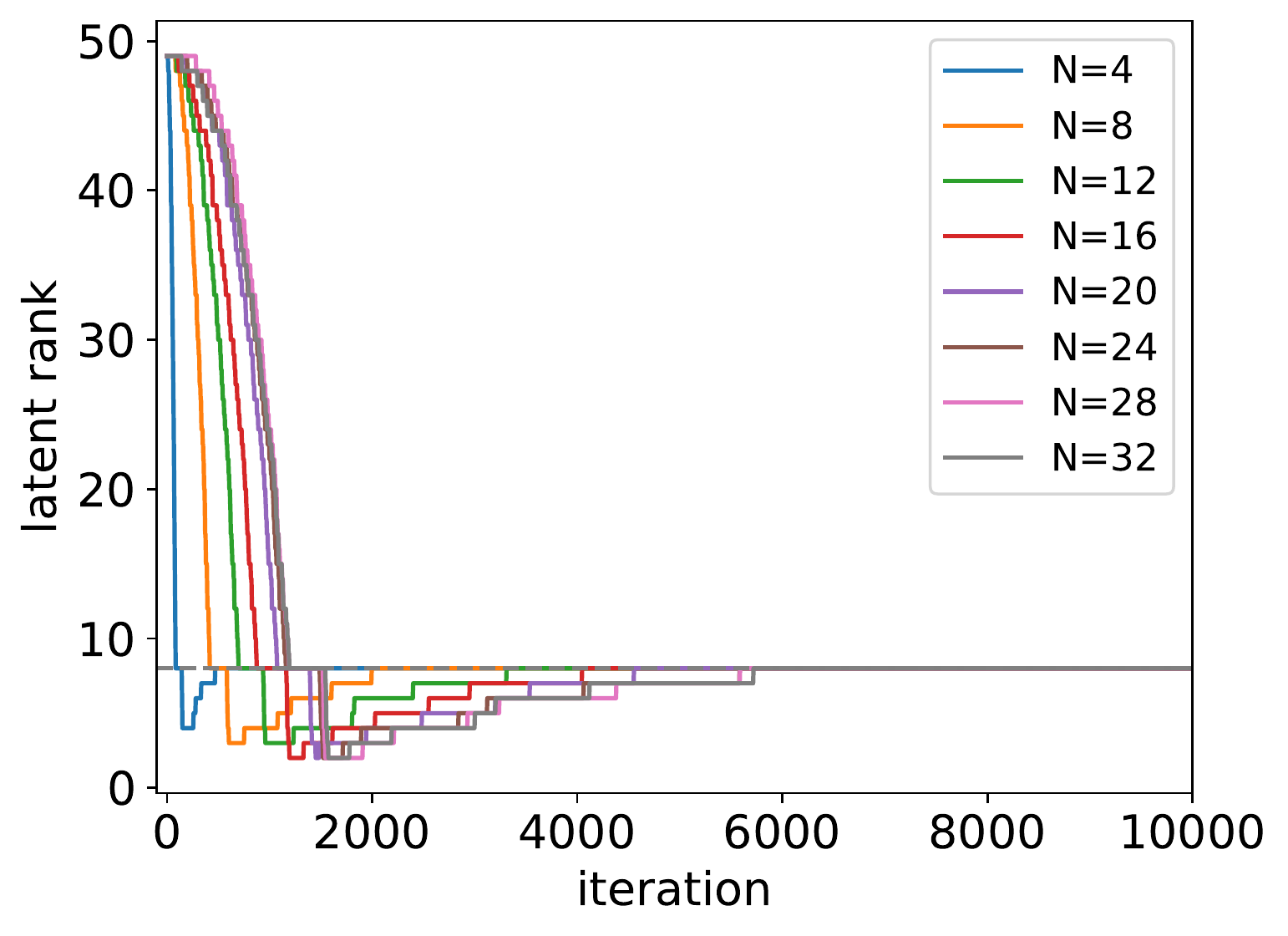}} {(d) orthogonal init ($\alpha=2.0$)}
\\
\end{tabular}
\caption{(a) vanilla AE: training loss (top) and latent code rank (bottom). (b) AE with linear sub-networks of varying depth $N$ with standard initialization: $W_\text{e}$ rank (top) and latent code rank (bottom). The dynamics are highly dependent on initial spectra of $W_\text{e}$ as a function of $N$. (c), (d) are the same setup as (b) but with scaled orthogonal initialization and learning rate adjustment (see Section~\ref{sec:orthogonal_init}, \ref{sec:lr_adjust} and \ref{sec:depth_norm}), where training behaviors are more stable with varying $N$.}
\vspace{-6pt}
\label{fig:linear_ae}
\end{figure*}

\subsection{Triggering Greedy Learning by Larger Updates}
\label{sec:lr_adjust}
On the surface, greedy rank learning would not emerge with a uniform spectrum from orthogonal initialization: in Equation~\ref{eq:update}, if $\mu_{r,r'}$ are the same with all components and $\frac{dL}{d\bold{z}_N}^\top \bold{z}^\top$ is distributed nearly uniformly across them, all components will be updated at similar rates. It is indeed possible to be stuck in this regime when $W_\text{e}$ is updated slowly compared to the rest of AE. From this non-greedy regime, however, greedy learning could be triggered by larger updates to $W_\text{e}$, which amplify modulation effect by increasing spectral heterogeneity. 

We increase magnitudes of weight updates to $W_\text{e}$ by reparameterizing all $W_i$ by $\alpha \hat{W}_i$: $\alpha$ is a constant ($\alpha > 1$) and $\hat{W}_i$ is learnable weights initialized by orthogonal matrices scaled by $1/\alpha$. In principle, this reparameterization by $\alpha$ has identical effects to scaling the local learning rate on $W_i$ by $\alpha^2$, but the reparameterization was found to be more numerically stable due to increased gradient magnitudes, especially with adaptive optimization like Adam \citep{adam}.

\subsection{Training Stability to Linear Depth $N$}
\label{sec:depth_norm}
Equation~\ref{eq:update} shows that weight updates depend on $\mu_{r,r'}$ as a function of $N$ ( Equation~\ref{eq:mu}). Assuming descending singular values, the dominant $\mu_{r,r'}$ is given by $r=r'=1$:
\begin{align}
\mu_{1,1} &= N \sigma_{1}^{2\frac{N-1}{N}} \nonumber
\end{align}
$\mu_{1,1}$ approaches $N \sigma_{1}^2$ with increasing $N$, so with a large $N$, learning speed of the dominant component is linear to $N$. Hence, we scale learning rate for the linear sub-network by $1/N$ to make its training speed more stable to $N$.

\section{Experiments}
\label{sec:experiments}
\subsection{Linear Autoencoder on Synthetic Data}
\label{sec:linear_ae}

In this section, we examine influence of $W_i$ initialization on training a simple linear AE. A 1D training dataset $X$ with $n$ samples, dimension $D$, and a known rank $r$ was synthesized by random Gaussian matrices $X_1 \in \mathbb{R}^{n \times r}$ and $X_2 \in \mathbb{R}^{r \times D}$:
\begin{align*}
X = X_1 X_2 + \epsilon
\end{align*}
where $\epsilon \in \mathbb{R}^{n \times D}$ denotes zero-mean Gaussian random noise with standard deviation $s$. In this linear case, the encoder and decoder are both a single matrix: $enc(\cdot) \in \mathbb{R}^{d \times D}$ and $dec(\cdot) \in \mathbb{R}^{D \times d}$. These parameters were set as follows: $n=64, D=256, d=128, r=8, s=0.2$. Note that this setup is under-determined with multiple solutions, but solutions that lead to latent codes of rank 8 would reflect the ground-truth data structure and overfit less to noise. See Appendix~\ref{appendix:linear_ae} for further experimental details.

Figure~\ref{fig:linear_ae}(a) shows vanilla linear AE without the linear sub-network. Without regularization, rank of latent codes stays high throughout training. In our experiments, the rank is defined as the number of singular values that exceed 0.01 after normalization by the largest one.

Figure~\ref{fig:linear_ae}(b) shows linear AE with a linear sub-network of varying depth $N$, initialized by standard He initialization \citep{He_init}. The evolution of $W_\text{e}$ rank and latent code rank highly depends on initial spectra induced by multiplication of the $N$ random matrices. For example, with $N=4$, the modulation effect is not pronounced and hence rank of $W_\text{e}$ stays high. On the contrary, with a too large $N$ (e.g. 24, 28, 32), the eigenvalues of $P_{W_\text{e}}$ are small and hence learning nearly stalls. Thus, to determine $N$ that leads to a good latent rank (i.e. $N$ within $[4, 16]$ in this experiment), one needs to rely on downstream tasks to assess quality of latent codes across varying values of $N$, for instance.

Our method avoids the initial spectral bias and hence leads to more stable rank evolution with varying $N$ as shown in Figure~\ref{fig:linear_ae}(c). The evolution is nearly invariant to $N$ for $N>20$. The effect of increasing $\alpha$ (Section \ref{sec:lr_adjust}) on training speed is shown in Figure~\ref{fig:linear_ae}(d).

\subsection{Nonlinear Autoencoder on MNIST}
\label{sec:nonlinear_ae}
We validated our method in nonlinear AE following experimental setups in \citep{irmae}. We trained an AE on the whole training set of MNIST \citep{mnist}, with a linear sub-network of varying depth $N$ at the bottleneck, initialized by either standard or orthogonal initialization. The latent code dimension was 128, and its quality was assessed by two downstream tasks: classification in the low-data regime and image sampling. All configurations in this experiment were run with 10 random seeds and the means are reported. See Appendix~\ref{appendix:nonlinear_ae_mnist} for experimental details.

\begin{figure}[t!]
\centering
\begin{tabular}{cc}
\subf{\includegraphics[width=38mm]{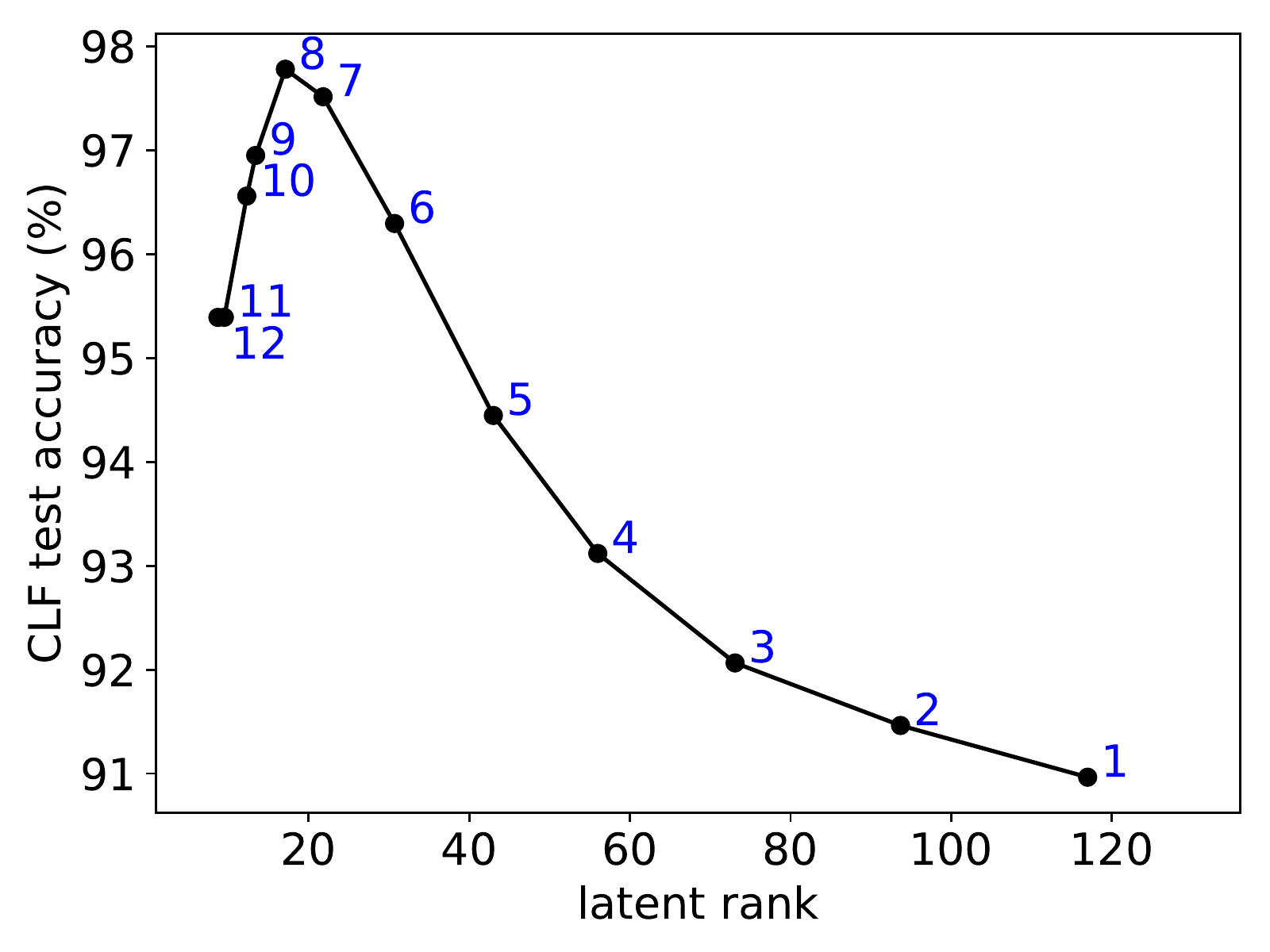}}{(a) classification} &
\subf{\includegraphics[width=38mm]{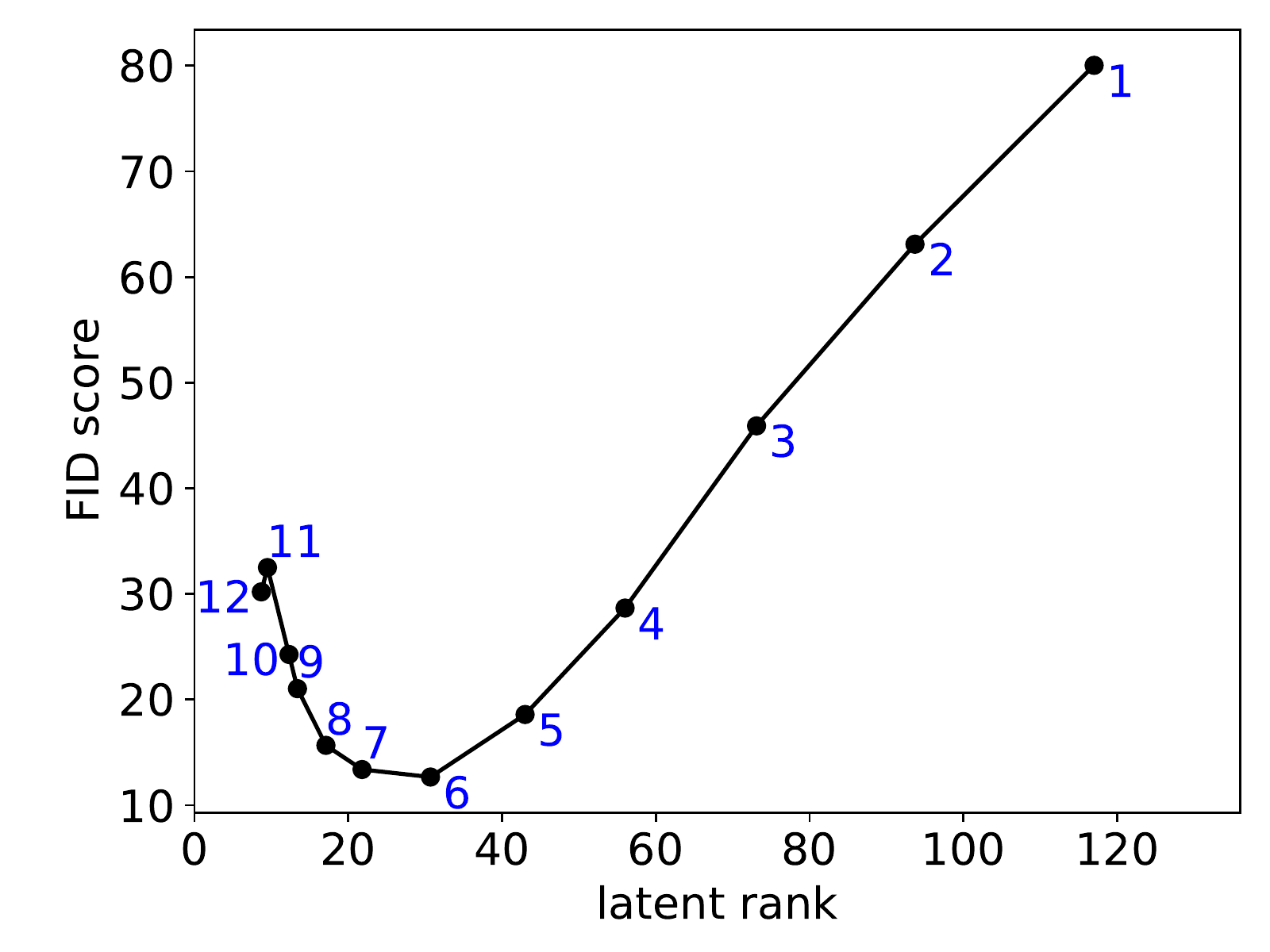}}{(b) sample quality}
\end{tabular}
\caption{(a) classification test accuracy and (b) sample quality in FID scores (lower is better) with varying latent ranks resulted from different depth $N$ (annotated in blue) with standard initialization.}
\label{fig:std_init_nonlinear}
\end{figure}

\begin{figure}[t!]
\centering
\begin{tabular}{cc}
\subf{\hspace{-4 pt}\includegraphics[width=39mm]{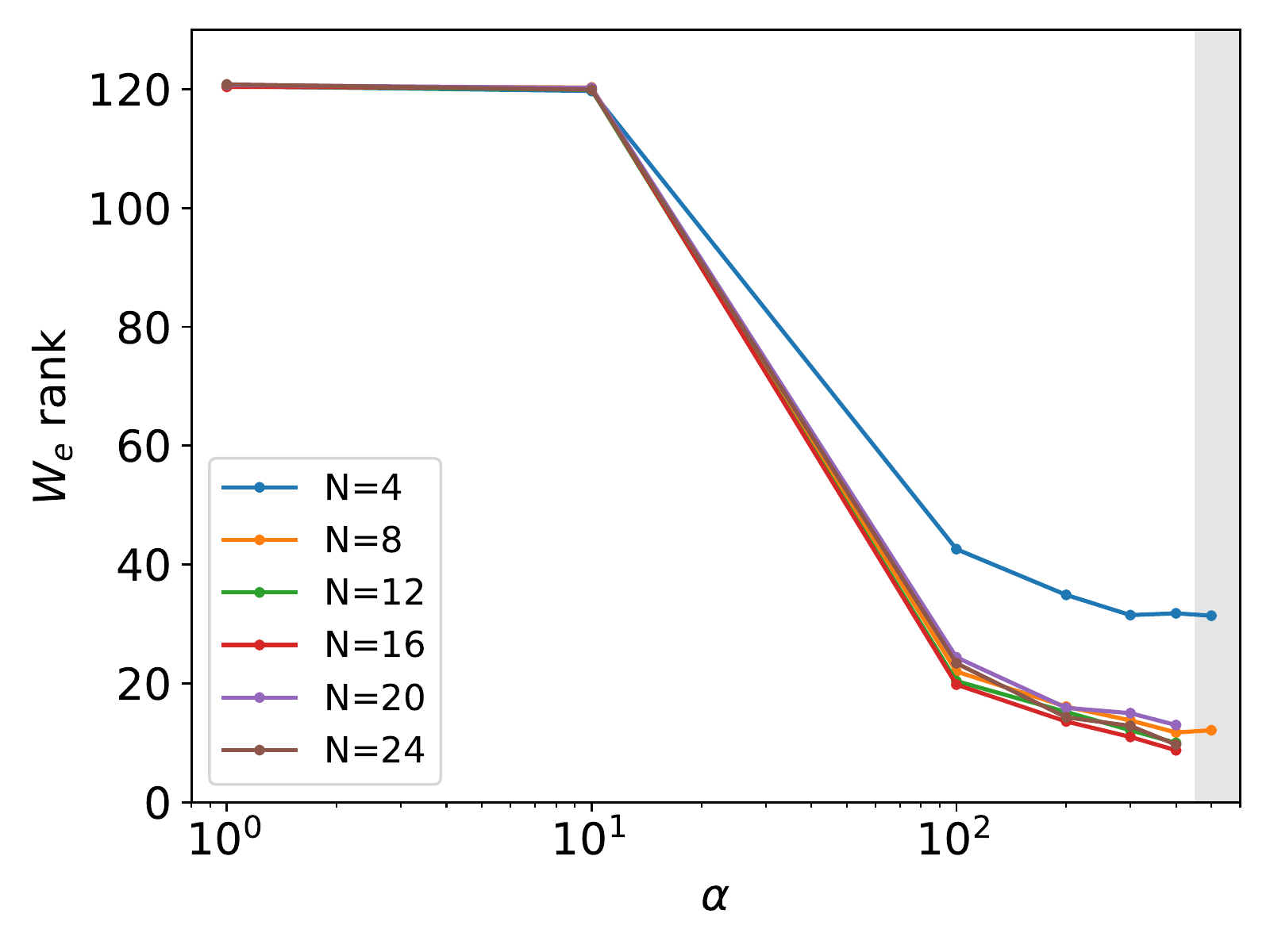}}{(a) $W_\text{e}$ rank} &
\subf{\hspace{-4 pt}\includegraphics[width=39mm]{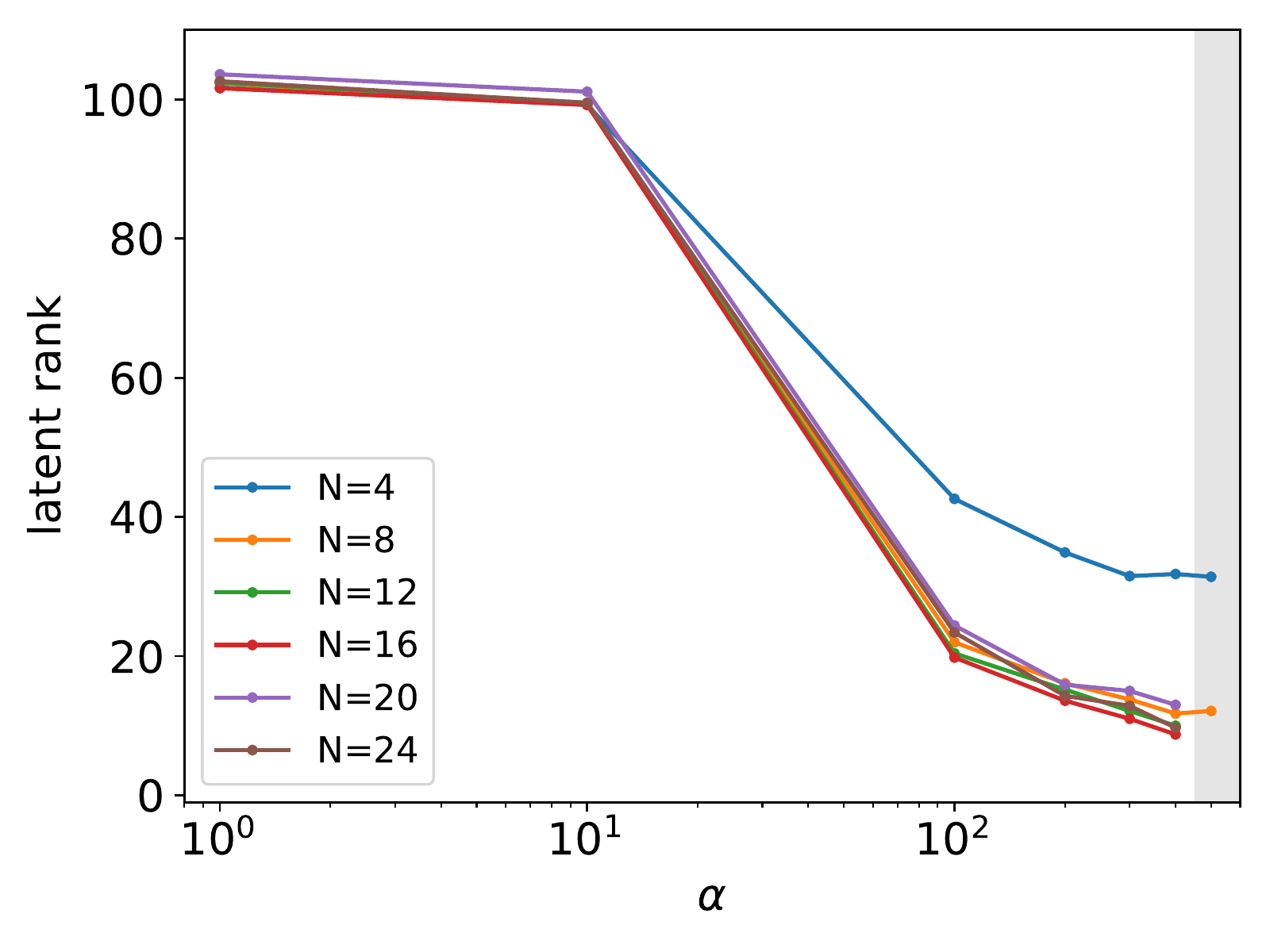}}{(b) latent rank} 
\\
\subf{\hspace{-4 pt}\includegraphics[width=39mm]{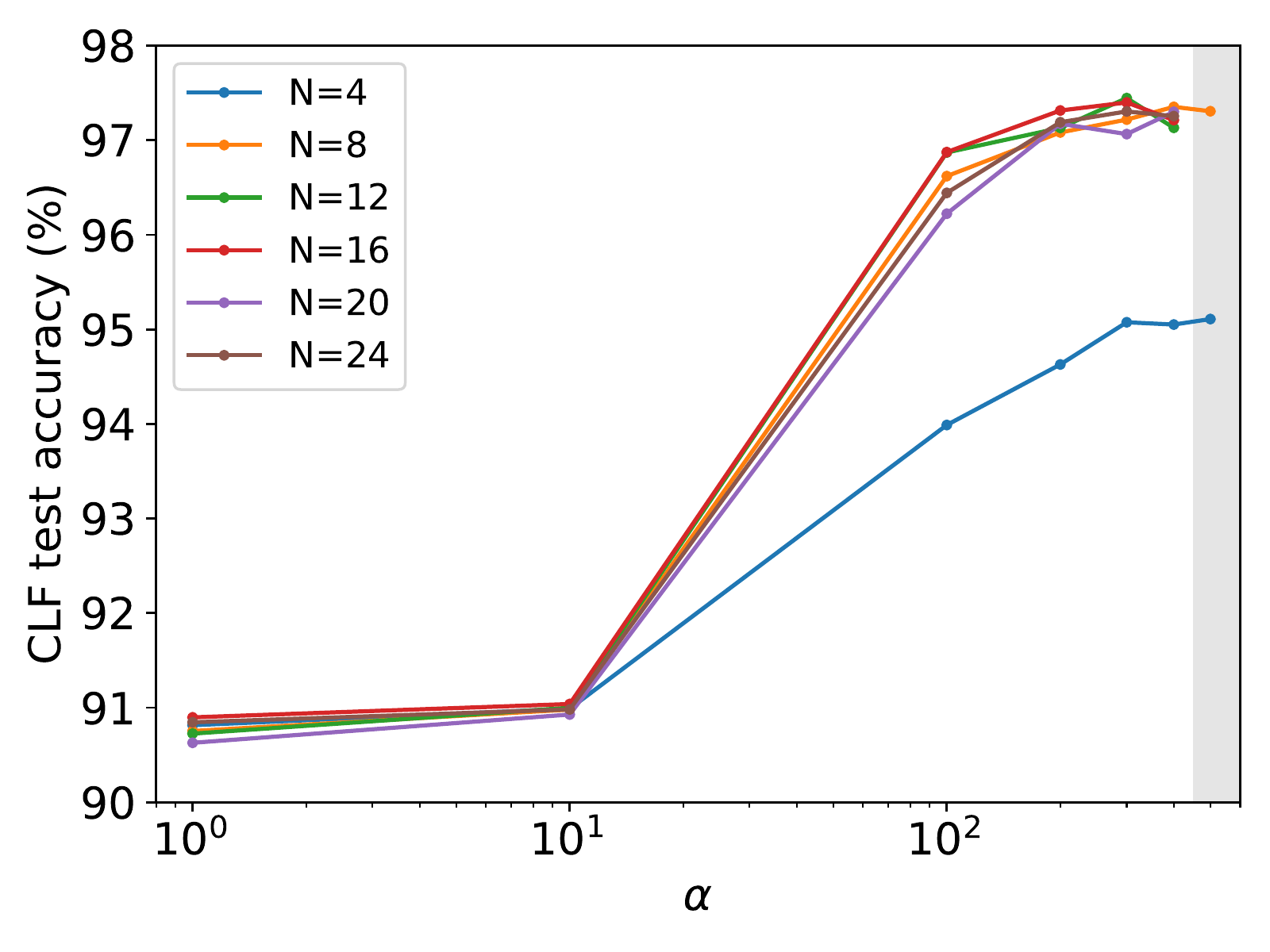}}{(c) classification} &
\subf{\hspace{-4 pt}\includegraphics[width=39mm]{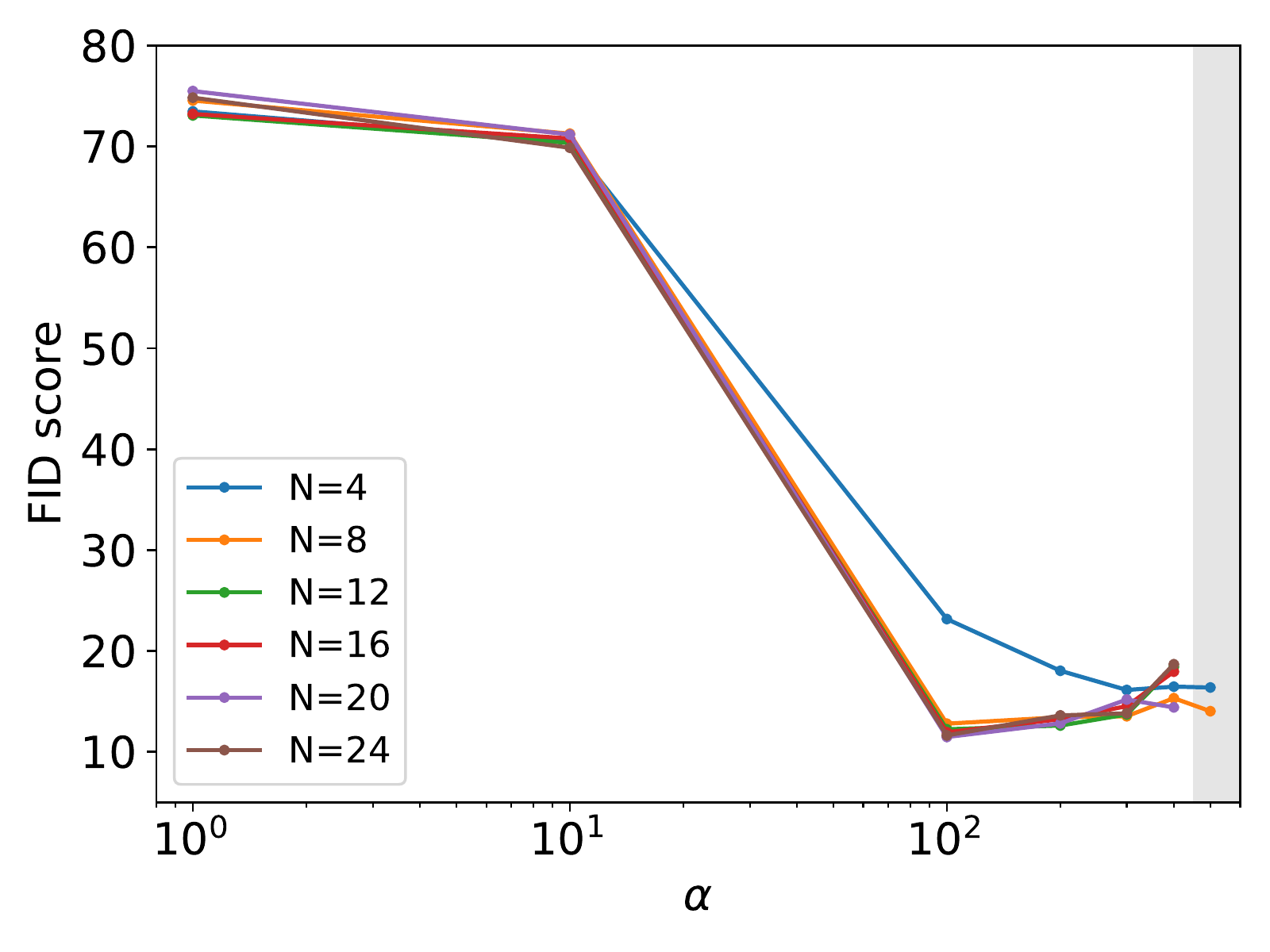}}{(c) sample quality}
\end{tabular}
\caption{With our method, model behaviors are influenced by training speed of $W_\text{e}$ as a function of $\alpha$ (Section~\ref{sec:lr_adjust}). The behaviors are similar for $N>8$. Shaded regions indicate $\alpha$ values that tend to lead to training divergence for larger $N$.}
\vspace{-10pt}
\label{fig:iro_nonlinear}
\end{figure}

\begin{figure}[t]
\centering
\begin{tabular}{cc}
\subf{\includegraphics[width=37mm]{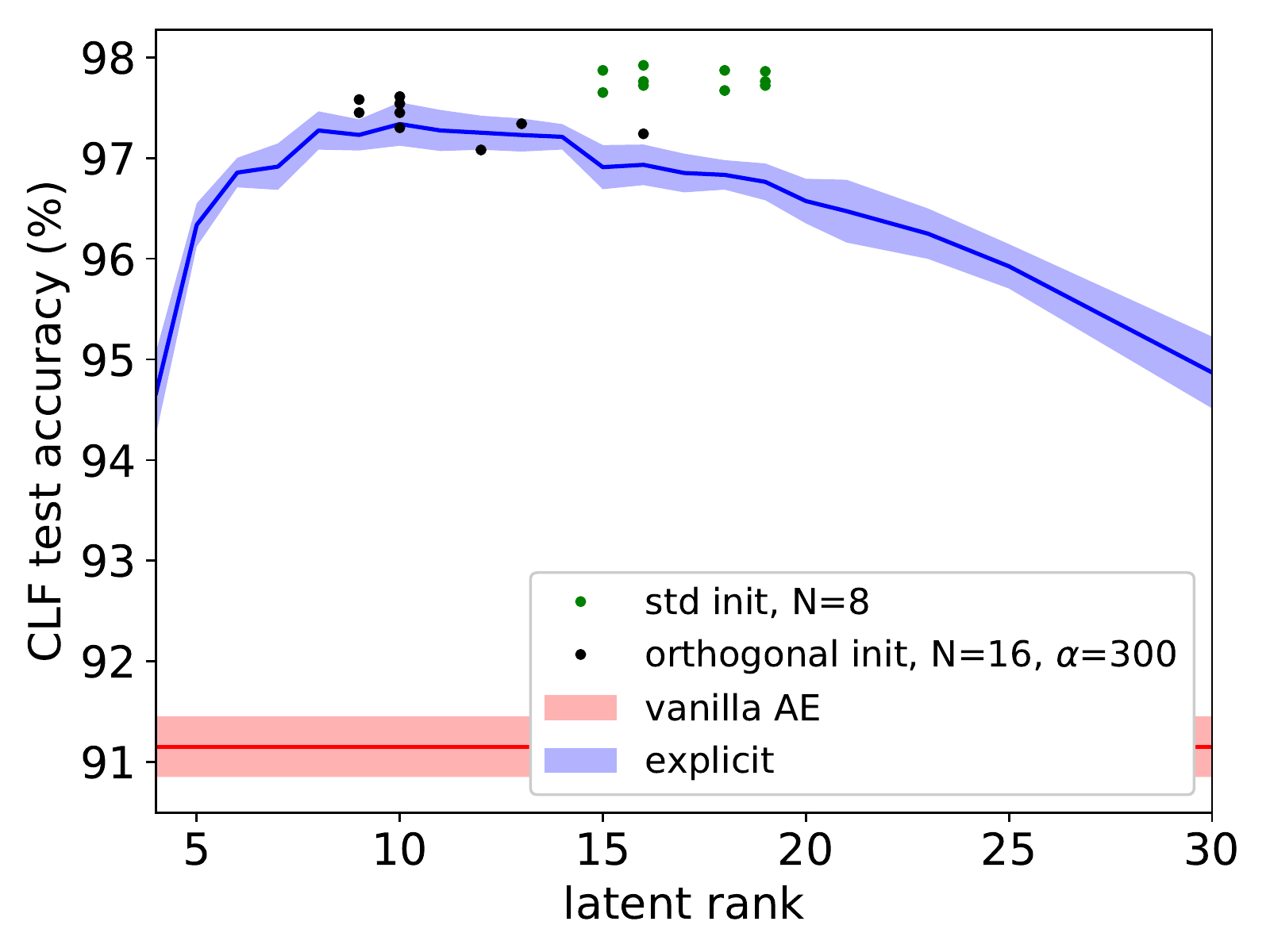}}{(a) classification} &
\subf{\includegraphics[width=37mm]{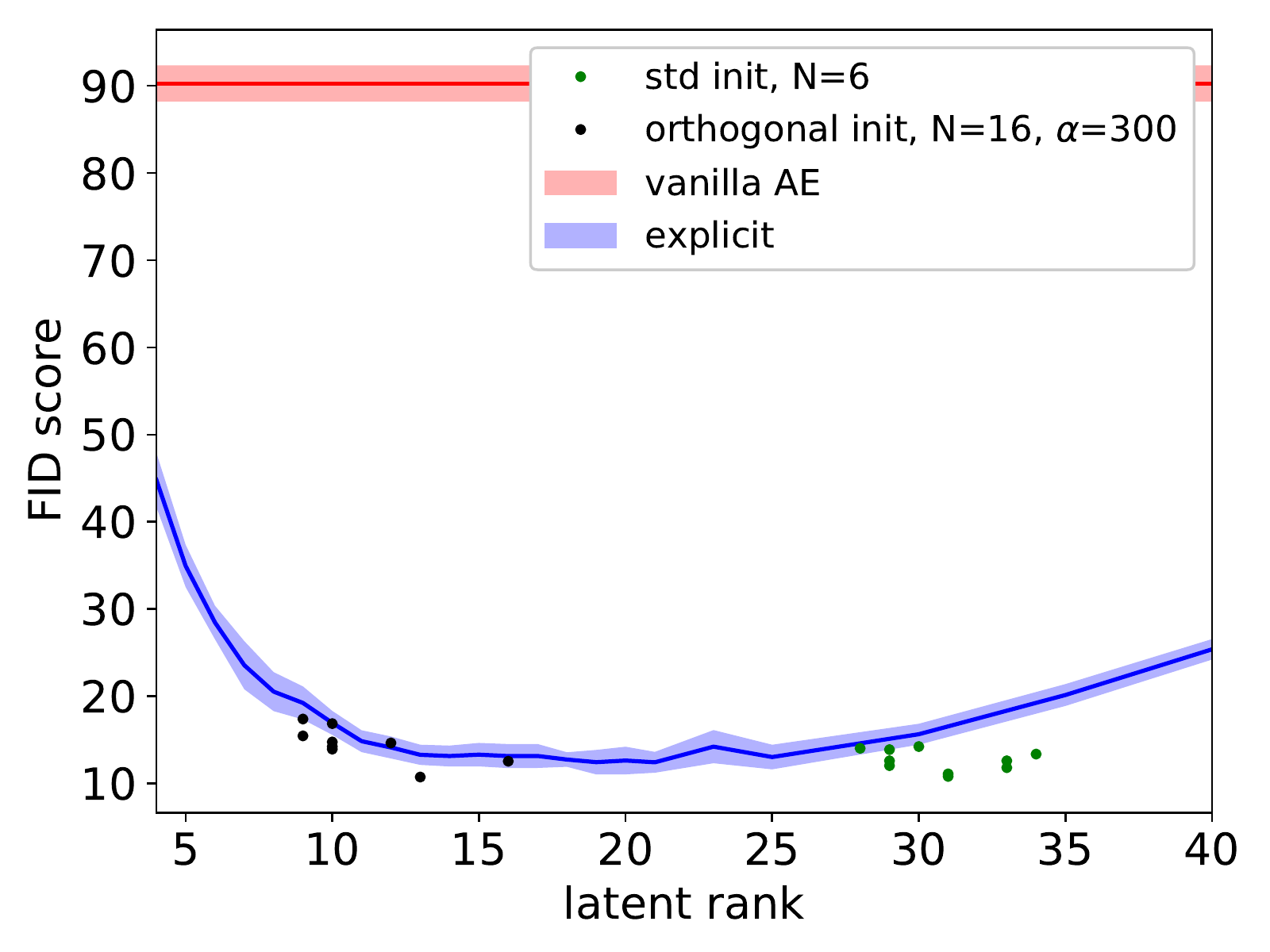}}{(b) sample quality}
\end{tabular}
\caption{Comparison of (a) classification test accuracy and (b) sample quality in FID scores (lower is better) at varying latent code ranks. Note that with vanilla AE, latent codes stay full-rank (i.e. 128). For standard and orthogonal initialization, results from 10 random seeds are shown separately, while means and standard deviations are shown for vanilla AE and the explicit method.}
\vspace{-10pt}
\label{fig:compare_nonlinear}
\end{figure}

As a baseline method for rank-regularized AE, we inserted a 2-layer linear sub-network with varying shared dimensions to explicitly control the latent rank, which is referred to as ``explicit'' in Figure~\ref{fig:compare_nonlinear}. This method already shows significant improvement over the vanilla AE in downstream tasks, but to determine the optimal latent rank, one needs to rely on downstream tasks to search over varying shared dimensions.

Our training method involved two stages: rank optimization with orthogonal initialization followed by loss minimization where $W_\text{e}$ is explicitly constrained to the estimated rank. Rank optimization first proceeded until $W_\text{e}$ rank no longer changed for 5 epochs. Training continued with a 2-layer linear sub-network with shared dimension equal to the $W_\text{e}$ rank as replacement for the $N$-layer linear sub-network to minimize loss.

Figure~\ref{fig:std_init_nonlinear} shows that with standard initialization, downstream task performance highly depends on selection of $N$, and the optimal $N$ differs between the two tasks, suggesting the need to rely on downstream tasks to select a good $N$.

In contrast, with our method, model behaviors are similar across varying $N$ for $N>8$, as shown in Figure~\ref{fig:iro_nonlinear}. Performance of both tasks improve with increasing learning speed of $W_\text{e}$ as a function of $\alpha$ (Section~\ref{sec:lr_adjust}) and converge approximately after $\alpha > 100$. A too large $\alpha$ could result in training divergence similar to a too large learning rate in typical training. In our experiments, we increased $\alpha$ in fixed increments of 100 until training divergence happened. 

Figure~\ref{fig:compare_nonlinear} shows comparison of these methods. For standard initialization, the best $N$ for each task is shown, while for our method with orthogonal initialization, a fixed $\alpha$ in the convergence range is chosen regardless of tasks. Without parameter search on the downstream tasks as in the explicit method and standard initialization, our method converges to a range of latent ranks optimal for the tasks and achieves similar performance, which suggests the potential for estimating intrinsic data dimensions in unsupervised settings without relying on specific downstream tasks.

\section{Conclusion and Future Work}
In this paper, we studied implicit regularization induced by deep linear networks at autoencoder bottlenecks, revealing that latent codes are biased towards low-rank structures through greedy learning. We further showed that orthogonal initialization removes prior spectral bias and significantly improves training stability across linear network depths when combined with principled learning speed adjustment.

These findings suggest potential for estimating intrinsic data structures in unsupervised settings, but some questions remain open. For example, it is still unclear whether there exist universally optimal latent ranks, since in some scenarios, different tasks could require different ranks to perform optimally. (One such scenario we observed on CIFAR-10 \citep{cifar} is discussed in Appendix~\ref{appendix:nonlinear_cifar10}.) Latent code optimality could also depend on capacities of the encoder, decoder, and models in downstream tasks, and hence interplay between these elements should be further studied.

\bibliography{reference}
\bibliographystyle{icml2021}

\appendix

\section{Greedy Rank Learning: Evolution of Top Singular Values}
\label{appendix:greedy}

\begin{figure}[h!]
\centering
\begin{tabular}{cc}
\subf{\hspace{-6pt} \includegraphics[width=40mm]{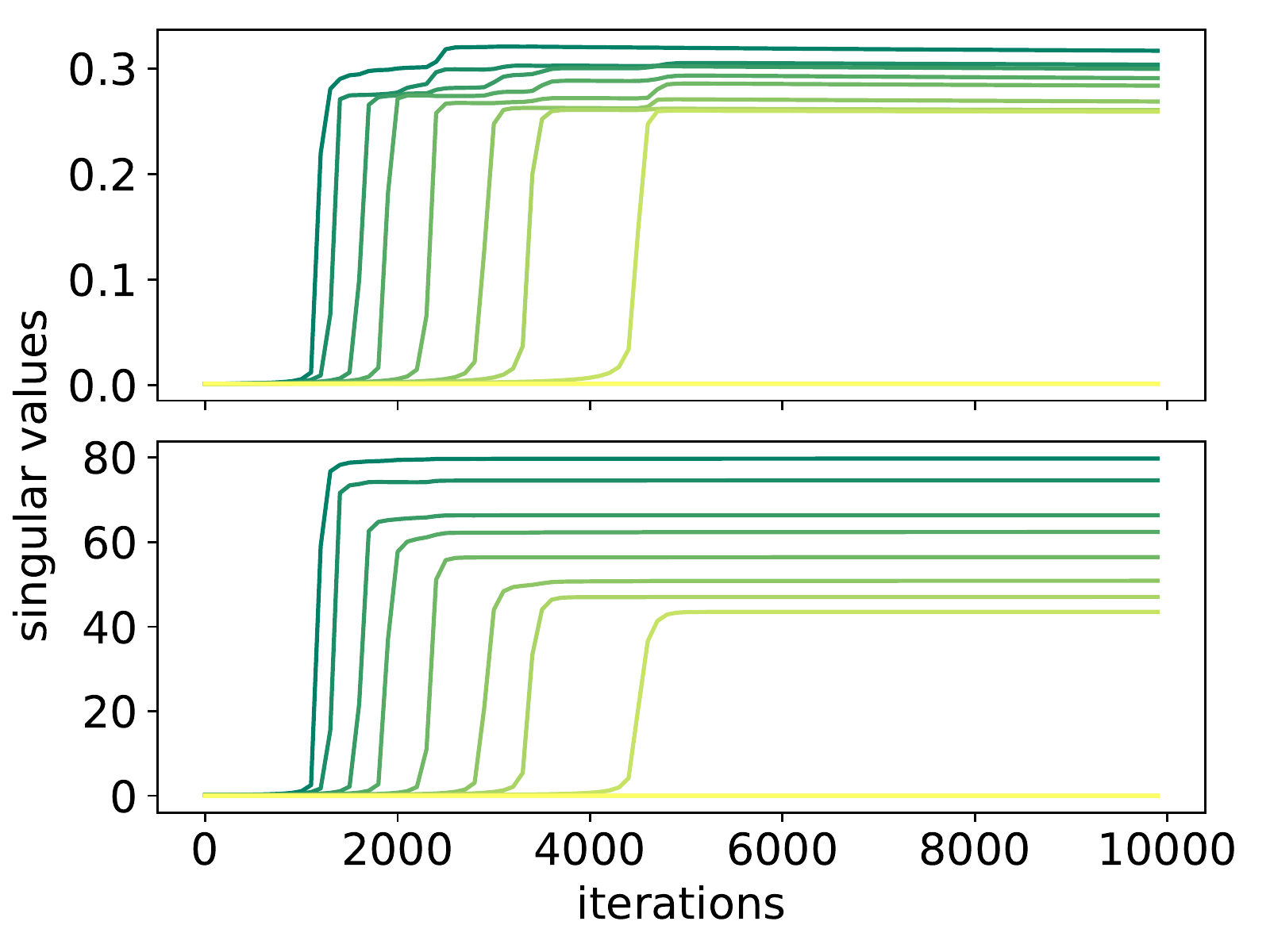}}{(a) linear AE} &
\subf{\hspace{-10pt}\includegraphics[width=40mm]{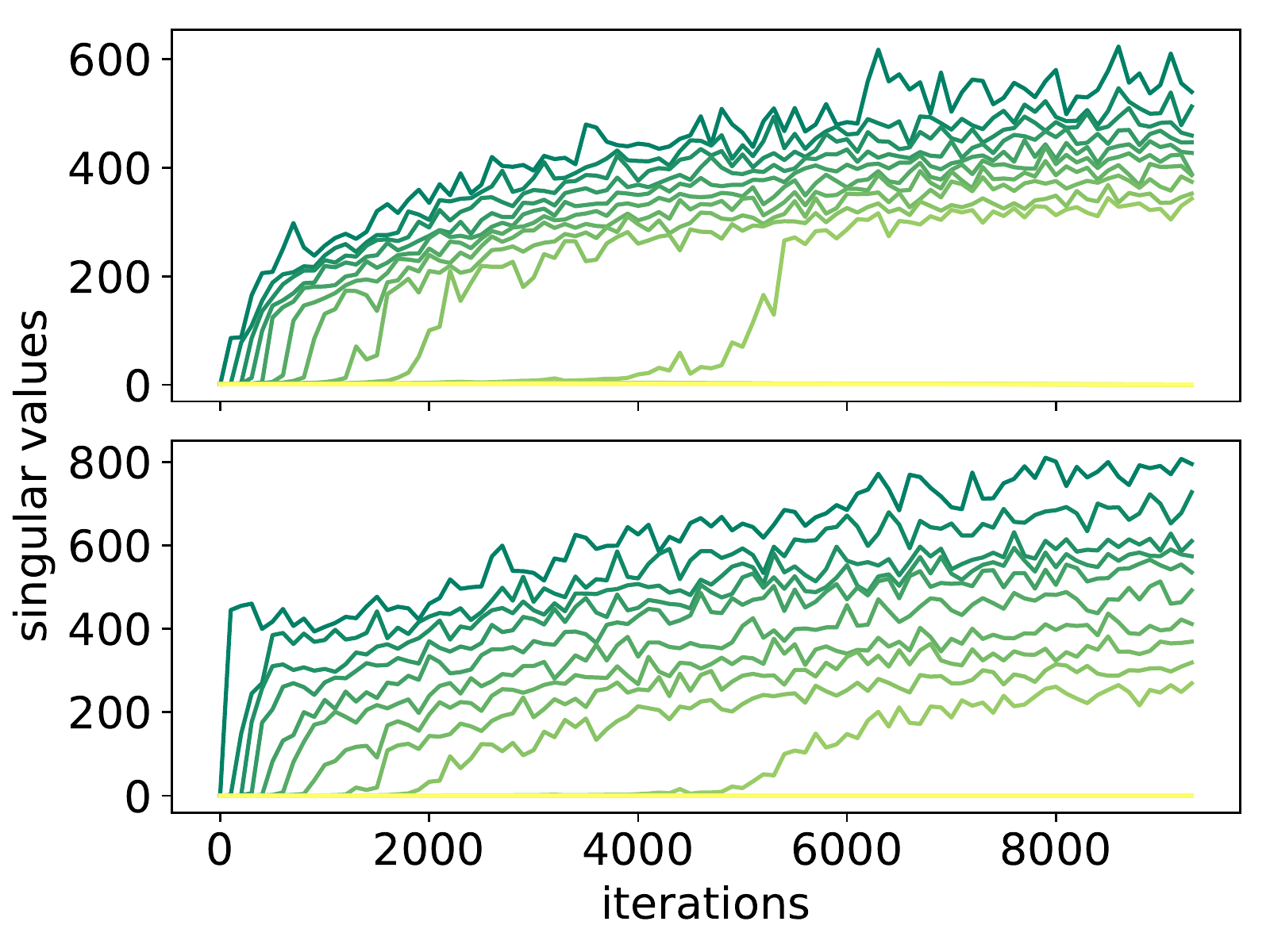}} {(b) nonlinear AE}
\end{tabular}
\caption{Evolution of top singular values of $W_\text{e}$ (top) and latent codes (bottom) with our method for (a) linear AE (Section \ref{sec:linear_ae}) and (b) nonlinear AE (Section \ref{sec:nonlinear_ae}).}
\label{fig:greedy}
\end{figure}

Figure~\ref{fig:greedy} shows representative evolution patterns of top singular values with our method to demonstrate the greedy learning behavior. For the linear AE in Figure~\ref{fig:greedy}(a), linear sub-network depth $N=16$ and $\alpha=2.0$. For the nonlinear AE in Figure~\ref{fig:greedy}(b), $N=16$ and $\alpha=300$.

\section{Linear Autoencoder on Synthetic Data: Experimental Details}
\label{appendix:linear_ae}
In this experiment, the encoder and decoder were single matrices initialized by random zero-mean Gaussian weights of a 0.1 standard deviation, and no bias was used. Weights were optimized using MSE loss by batch gradient descent with a learning rate of 0.03 for 10000 steps. For standard initialization, the linear sub-network was initialized by the standard He initialization \citep{He_init}. For our method, each linear layer $W_i$ was initialized by an independent random orthogonal matrix \citep{orthogonal_init}, scaled by a constant factor $0.001^{1/N}$ so that $W_\text{e}$ was effectively scaled by 0.001. Training was run in 64-bit floating-point precision.

\section{Nonlinear Autoencoder on MNIST: Experimental Details}
\label{appendix:nonlinear_ae_mnist}
In this experiment, we used the convolutional encoder and decoder architectures described in the MNIST experiments of \citep{irmae}. All images were resized to $32 \times 32$. The AE was trained for 200 epochs by using the Adam optimizer on the MSE loss with a learning rate of 0.0001 and a batch size of 32. Training was run in 64-bit floating-point precision.

For downstream classification, we froze the AE and trained a 2-layer ReLU MLP classifier on top of the latent codes using 1000 training samples. The classifier was trained for 500 epochs using the Adam optimizer on the cross-entropy loss, with a learning rate of 0.001 and batch size of 32. Best test accuracy on the test set throughout training is reported.

For image sampling, we fit a Gaussian mixture model with $k=4$ on latent codes of the training set, and generated images by running 10000 sampled codes through the decoder. Sample quality was measured by the Fr\'{e}chet Inception Distance (FID) \citep{fid_score, fid_pytorch} against the training set, where a lower score indicates better alignment of distributions between samples and training images.

\section{Nonlinear Autoencoder on CIFAR-10}
\label{appendix:nonlinear_cifar10}
Following MNIST experiments in Section~\ref{sec:nonlinear_ae}, we further examined latent rank regularization for CIFAR-10 via deep linear sub-networks at bottlenecks of nonlinear AE, as well as explicit rank regularization. CIFAR-10 experimental setups were identical to those for MNIST, including AE architectures (except for changes to accommodate RGB image channels).

\begin{figure}[h]
\centering
\begin{tabular}{cc}
\subf{\includegraphics[width=38mm]{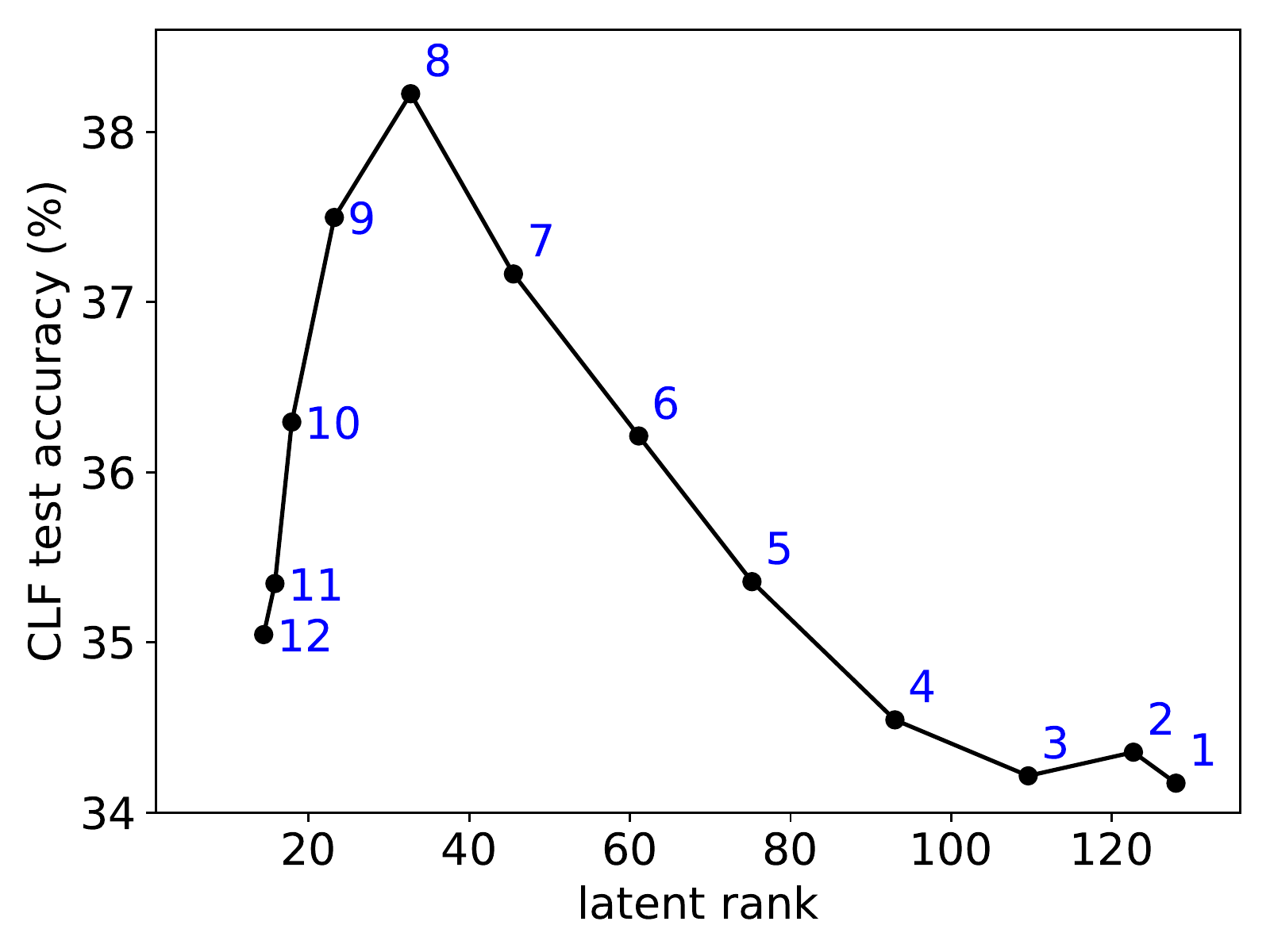}}{(a) classification} &
\subf{\includegraphics[width=38mm]{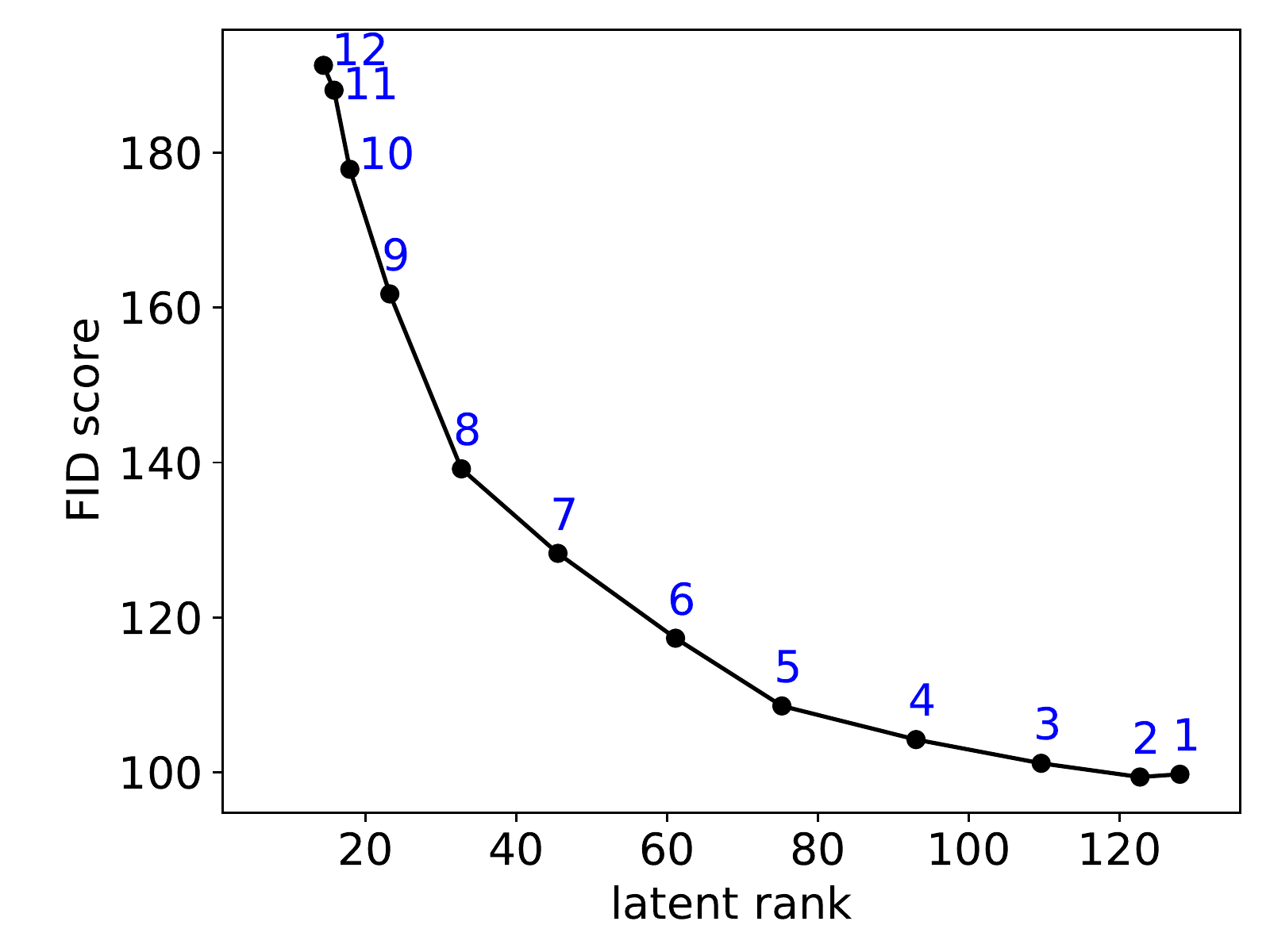}}{(b) sample quality}
\end{tabular}
\caption{CIFAR-10: (a) classification test accuracy and (b) sample quality in FID scores (lower is better) with varying latent ranks resulted from different depth $N$ (annotated in blue) with standard initialization. Note that the optimal rank for classification is around 32 while that for sample quality is 128 (full-rank).}
\label{fig:std_init_nonlinear_cifar10}
\end{figure}

\begin{figure}[h]
\centering
\begin{tabular}{cc}
\subf{\hspace{-4 pt}\includegraphics[width=39mm]{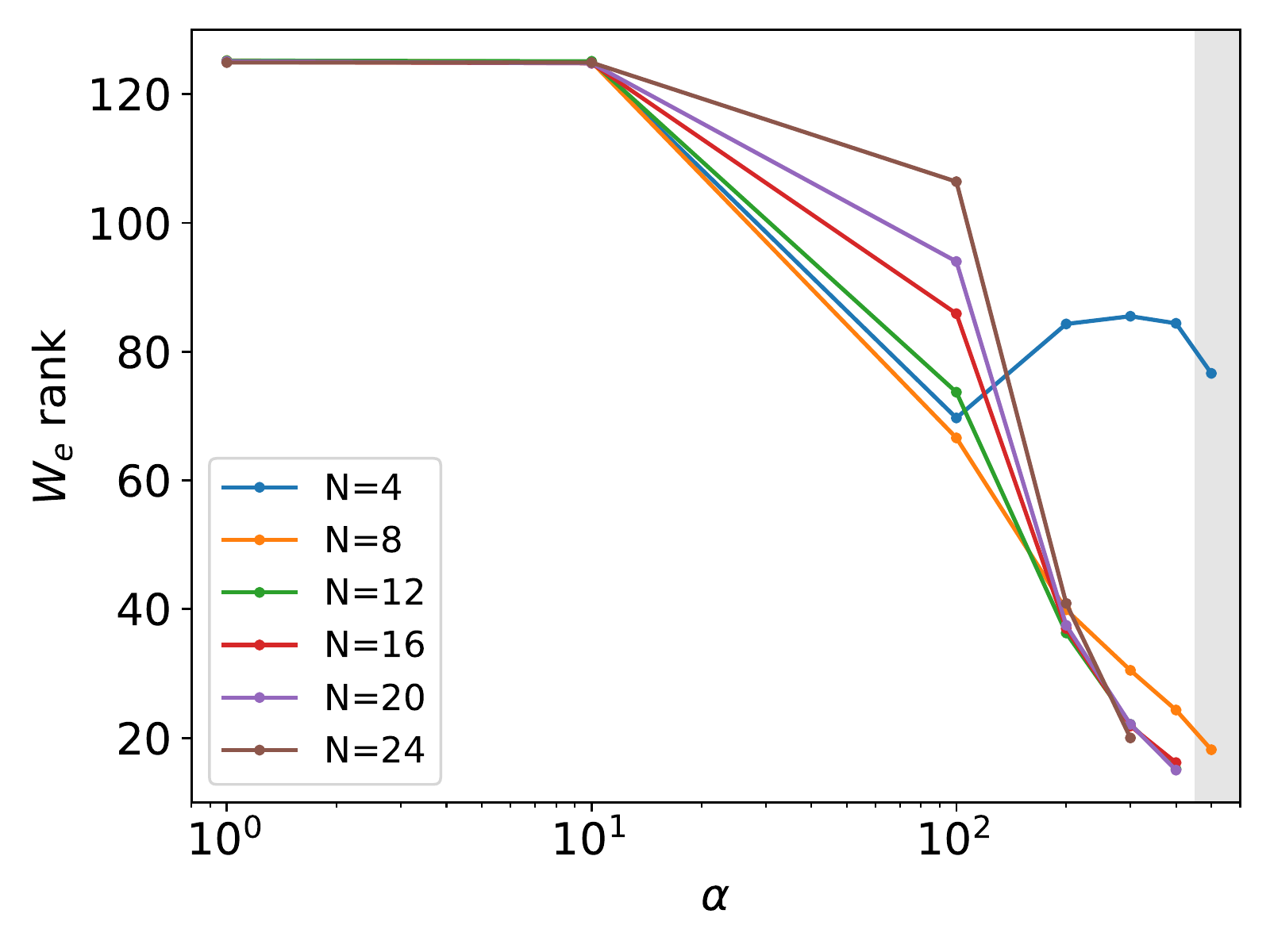}}{(a) $W_\text{e}$ rank} &
\subf{\hspace{-4 pt}\includegraphics[width=39mm]{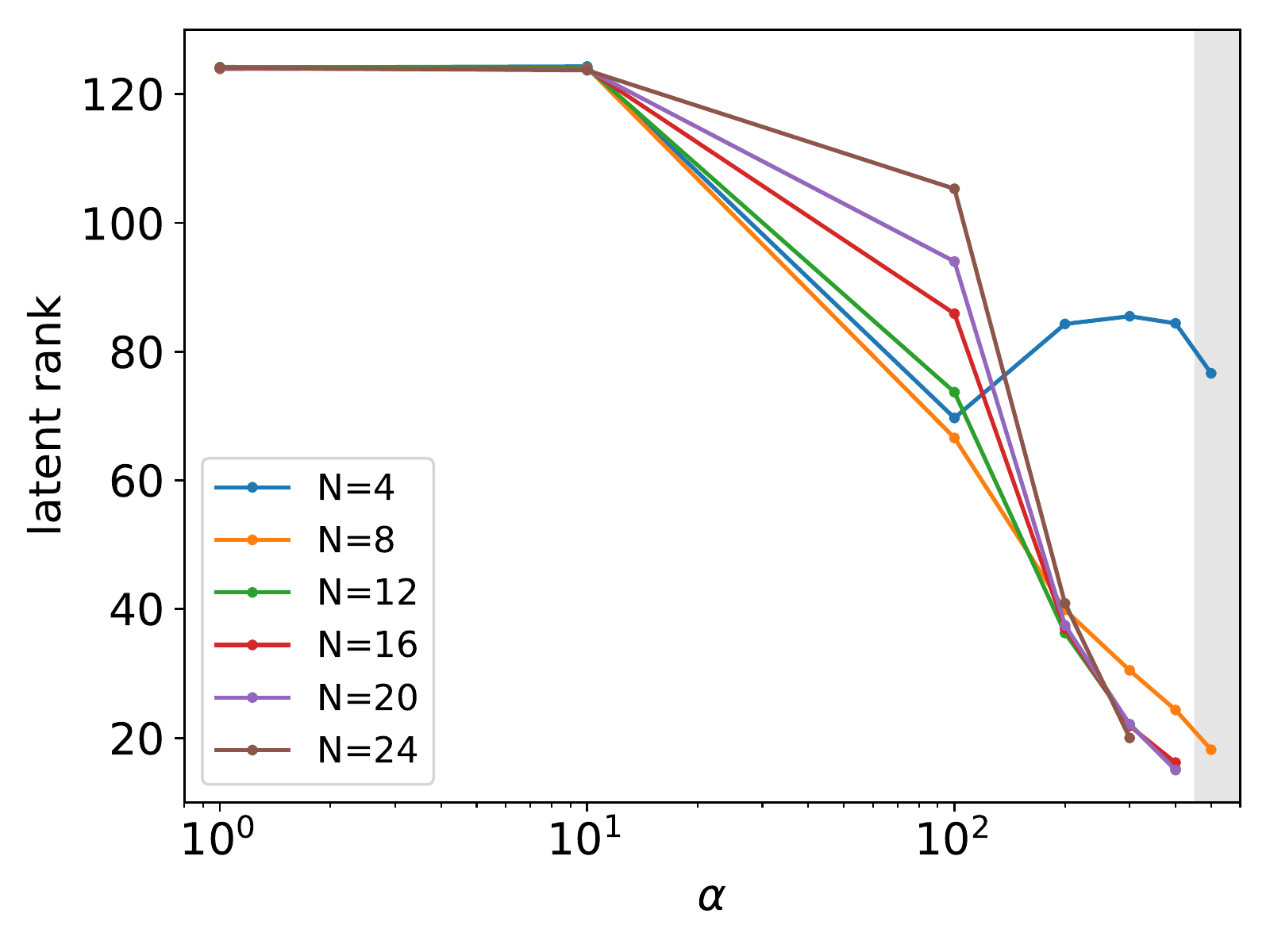}}{(b) latent rank} 
\\
\subf{\hspace{-4 pt}\includegraphics[width=39mm]{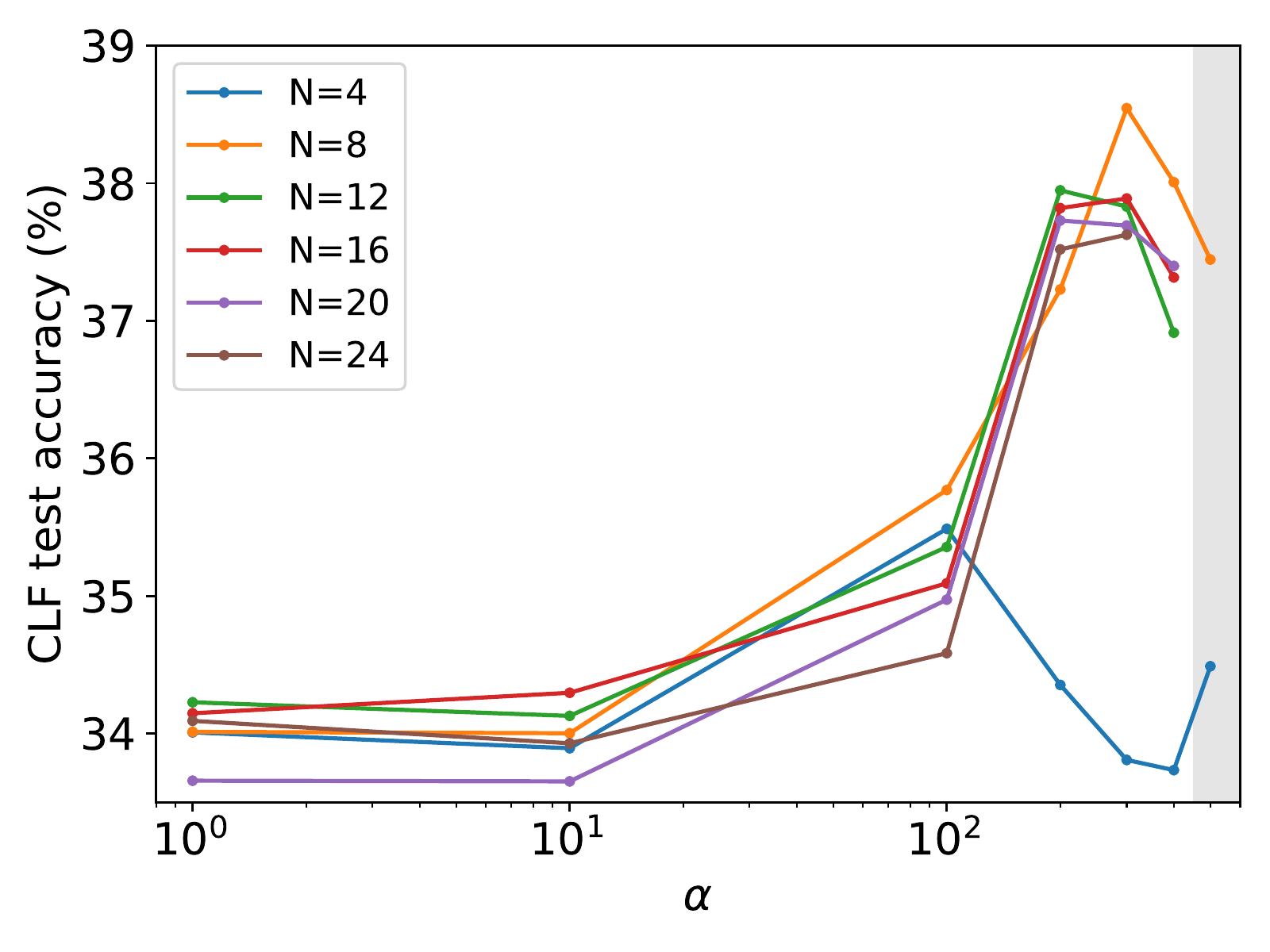}}{(c) classification} &
\subf{\hspace{-4 pt}\includegraphics[width=39mm]{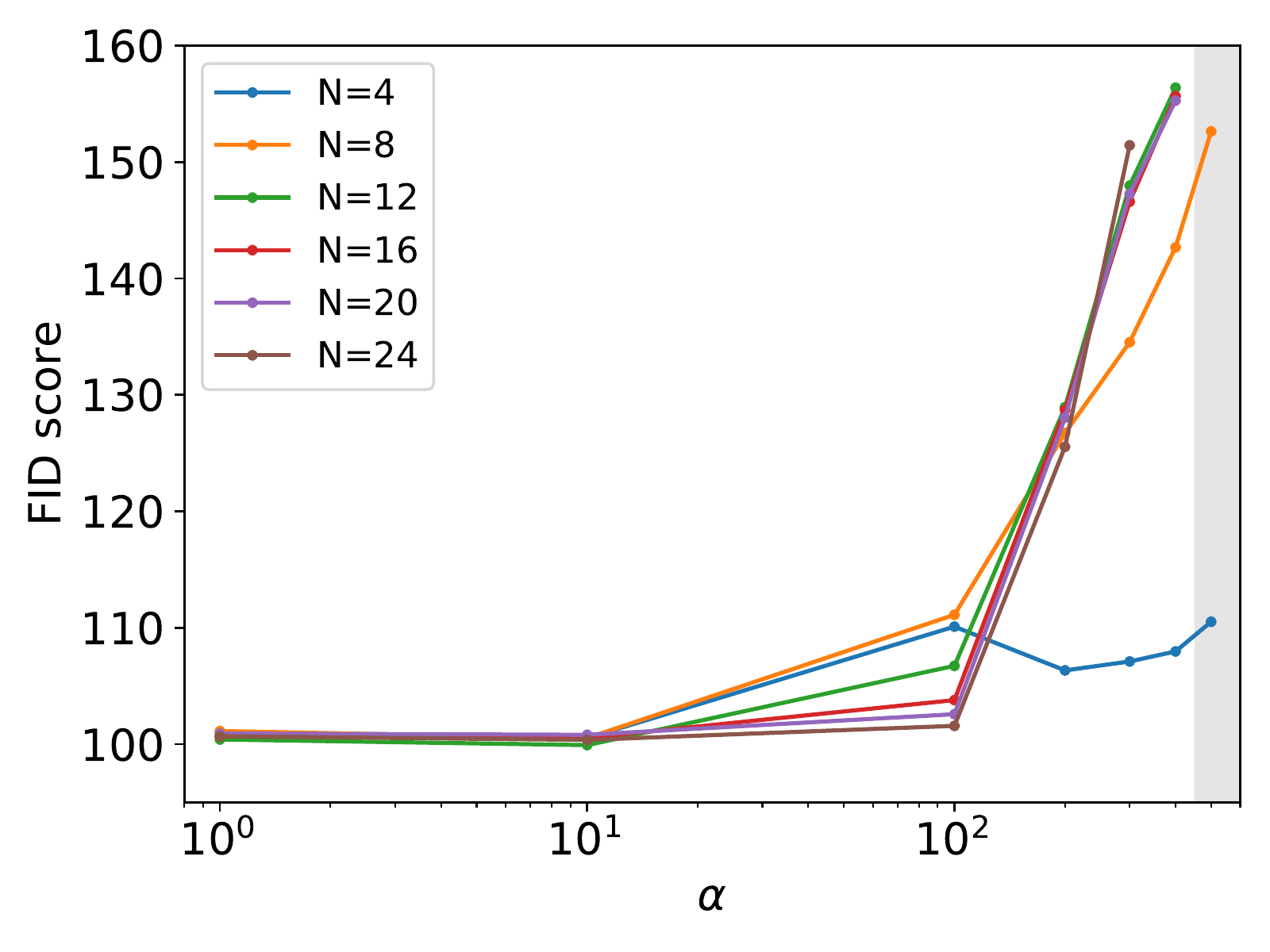}}{(c) sample quality}
\end{tabular}
\caption{CIFAR-10 with orthogonal initialization at varying $\alpha$ and depth $N$: with increasing $\alpha$, the latent rank decreases, which leads to improved classification test accuracy but worse FID scores (the lower the better). Shaded regions indicate $\alpha$ values that tend to lead to training divergence for larger $N$.}
\label{fig:iro_nonlinear_cifar10}
\end{figure}

\begin{figure}[h]
\centering
\begin{tabular}{cc}
\subf{\includegraphics[width=37mm]{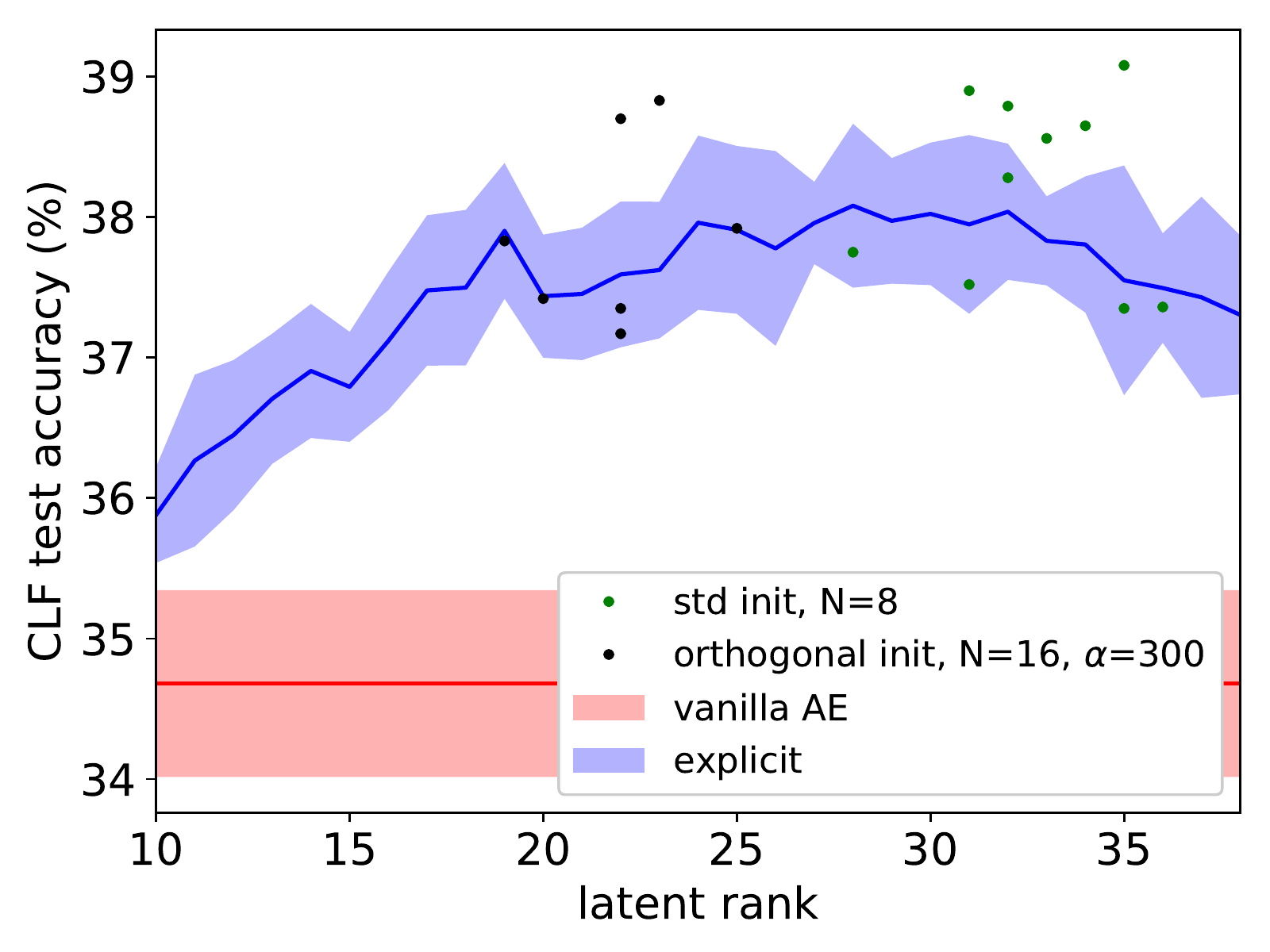}}{(a) classification} &
\subf{\includegraphics[width=37mm]{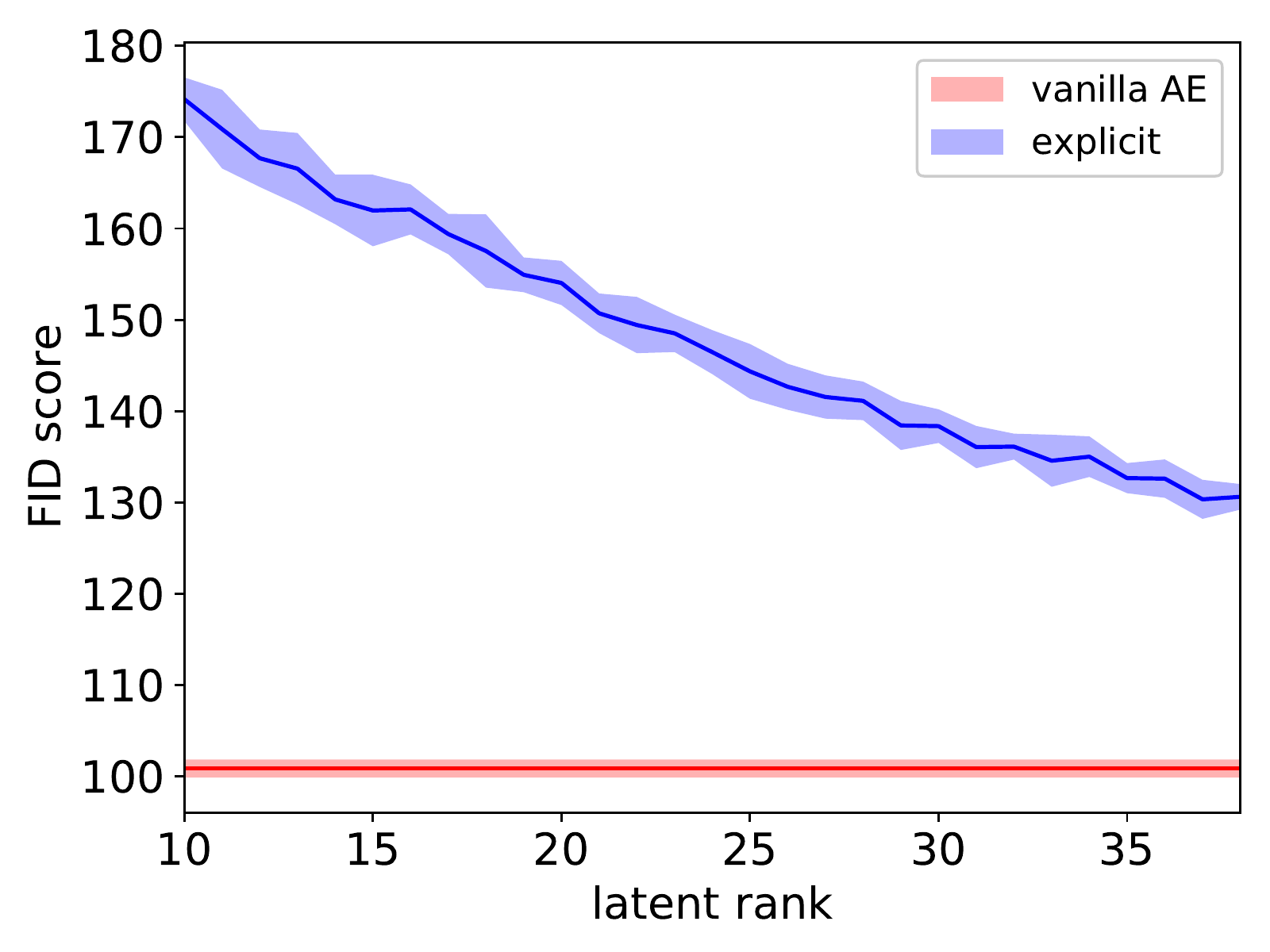}}{(b) sample quality}
\end{tabular}
\caption{CIFAR-10: comparison of (a) classification test accuracy and (b) sample quality in FID scores (lower is better) at varying latent code ranks. Note that with vanilla AE, latent codes stay full-rank (i.e. 128), which result in worse classification performance but better sample quality than lower-rank codes.}
\label{fig:compare_nonlinear_cifar10}
\end{figure}

It was found that in this scenario, optimal latent ranks could differ significantly between downstream tasks: in classification, optimal latent ranks were between 20 and 35 as indicated by explicit rank regularization, while full-rank codes (i.e. rank 128) performed the best in image sampling as measured by FID scores. The results are shown in Figure~\ref{fig:std_init_nonlinear_cifar10}, \ref{fig:iro_nonlinear_cifar10} and \ref{fig:compare_nonlinear_cifar10}.

This observation reveals critical questions to investigate in representation learning. One such question is whether there exist codes that perform optimally across downstream tasks, provided that downstream task models are designed properly to make use of the codes (e.g. class-conditional image sampling to model distributions of more sophisticated datasets like CIFAR-10). Another question concerns latent structures of different datasets, which is motivated by comparison between MNIST and CIFAR-10 with our method: while in MNIST, classification test accuracy plateaus with increasing $\alpha$ before divergence (Figure~\ref{fig:iro_nonlinear}(c)), the accuracy slightly drops at large $\alpha$ in CIFAR-10 as shown in Figure~\ref{fig:iro_nonlinear_cifar10}(c). Our hypothesis is that as a more sophisticated dataset, CIFAR-10 requires higher latent ranks to fit, and hence could require more training epochs, more sophisticated pre-processing, or more careful adjustment of the training dynamics. We leave these investigations for future work. 

\end{document}